\pdfoutput=1
\documentclass[conference, 11pt]{IEEEtran}
\IEEEoverridecommandlockouts
\usepackage[dvipsnames]{xcolor}

\usepackage[square,sort&compress,comma,numbers]{natbib}
\usepackage[utf8]{inputenc} 
\usepackage[T1]{fontenc}    
\usepackage{url}            
\usepackage{booktabs}       
\usepackage{amsfonts}       
\usepackage{nicefrac}       
\usepackage{microtype}      
\usepackage{times}
\usepackage{soul}
\usepackage[utf8]{inputenc}
\usepackage{graphicx}
\usepackage{booktabs}
\usepackage{outlines}
\usepackage{adjustbox}

\usepackage[colorlinks=true,
citecolor=blue,
filecolor=black,
linkcolor=blue,
urlcolor=blue]{hyperref}

\usepackage{microtype}
\usepackage{graphicx}
\usepackage{subfig}
\usepackage{booktabs} 

\usepackage{multirow}
\usepackage{paralist}

\usepackage{booktabs} 
\usepackage{multicol}
\usepackage{diagbox}

\usepackage{here}

\usepackage{amsmath,amssymb,amsfonts,amsbsy,amsfonts,latexsym}
\usepackage{multirow}
\usepackage{makecell}
\usepackage[labelfont=bf,textfont=it,belowskip=0pt,aboveskip=5pt,tableposition=top]{caption}
\usepackage{colortbl}

\definecolor{colorA}{RGB}{189,201,225}
\definecolor{colorB}{RGB}{103,169,207}
\definecolor{colorC}{RGB}{ 28,144,153}
\definecolor{colorD}{RGB}{  1,108, 89}

\newcolumntype{R}{>{\columncolor{gray!40}}r}
\newcolumntype{L}{>{\columncolor{gray!40}}l}
\newcolumntype{C}{>{\columncolor{gray!40}}c}

\usepackage{tabularx,colortbl}
\usepackage{multirow}
\usepackage[normalem]{ulem}
\useunder{\uline}{\ul}{}

\usepackage{enumitem}

\usepackage{xparse}

\DeclareGraphicsExtensions{.pdf,.png}

\usepackage{longtable}
\usepackage{pgfplots}

\usepackage{amsmath}
\usepackage{pifont}

\usepackage{booktabs}       
\usepackage{amsfonts}       
\usepackage{nicefrac}       
\usepackage{microtype}      
\usepackage{xspace}      

\usepackage[utf8]{inputenc} 
\usepackage[T1]{fontenc}    
\usepackage{chngcntr}

\usepackage[utf8]{inputenc}
\usepackage[english]{babel}

\newcommand{\loss}{\mathcal{L}}

\NewDocumentCommand{\var}{O{s} m O{}}{%
  \ensuremath{#1_{#2}^{#3}}
}
\usepackage{siunitx}

\definecolor{light-gray}{gray}{0.80}

\newcommand\eref{Eq.~\ref}
\newcommand\fref{Figure~\ref}

\newcommand\sref{Section~\ref}

\definecolor{brickred}{rgb}{0.8, 0.25, 0.33}
\definecolor{brickred2}{rgb}{0.25, 0.8, 0.33}

\begin{document}

\title{A Survey of Quantization Methods for Efficient Neural Network Inference}

\author{
\large
Amir Gholami$^{*}$\thanks{\noindent$^{*}$Equal contribution.},
Sehoon Kim$^{*}$,
Zhen Dong$^{*}$,
Zhewei Yao$^{*}$,
Michael W. Mahoney,
Kurt Keutzer\\
University of California, Berkeley\\
{\tt\small \{amirgh, sehoonkim, zhendong, zheweiy, mahoneymw, keutzer\}@berkeley.edu}
}
\maketitle
\pagestyle{plain}


\begin{abstract}
As soon as abstract mathematical computations were adapted to computation on digital computers, the problem of efficient representation, manipulation, and communication of the numerical values in those computations arose. 
Strongly related to the problem of numerical representation is the problem of quantization: in what manner should a set of continuous real-valued numbers be distributed over a fixed discrete set of numbers to minimize the number of bits required and also to maximize the accuracy of the attendant computations?  
This perennial problem of quantization is particularly relevant whenever memory and/or computational resources are severely restricted, and it has come to the forefront in recent years due to the remarkable performance of Neural Network models in computer vision, natural language processing, and related areas. 
Moving from floating-point representations to low-precision fixed integer values represented in four bits or less holds the potential to reduce the memory footprint and latency by a factor of 16x; and, in fact, reductions of 4x to 8x are often realized in practice in these applications.
Thus, it is not surprising that quantization has emerged recently as an important and very active sub-area of research in the efficient implementation of computations associated with Neural Networks. 
In this article, we survey approaches to the problem of quantizing the numerical values in deep Neural Network computations, covering the advantages/disadvantages of current methods.
With this survey and its organization, we hope to have presented a useful snapshot of the current research in quantization for Neural Networks and to have given an intelligent organization to ease the evaluation of future research in this area.
\end{abstract}


\section{Introduction}
\label{sec:intro}

Over the past decade, we have observed significant improvements in the
accuracy of Neural Networks (NNs) for a wide range of problems, often
achieved by highly over-parameterized models.
While the accuracy of these over-parameterized (and thus very large) NN models has significantly increased, the sheer size of these models means that it is not possible to deploy them for many resource-constrained applications.
This creates a problem for realizing pervasive deep learning, which requires real-time inference, with low energy consumption and high accuracy, in resource-constrained environments.
This pervasive deep learning is expected to have a significant impact on a wide range of applications such as
real-time intelligent healthcare monitoring, autonomous driving,
audio analytics, and speech~recognition.

Achieving efficient, real-time NNs with optimal accuracy requires rethinking the design, training, and deployment of NN models~\cite{gholami2020integrated}.
There is a large body of literature that has focused on addressing these issues by making NN models more efficient (in terms of latency, memory footprint, and energy consumption, etc.), while still providing optimal accuracy/generalization trade-offs. 
These efforts can be broadly categorized as follows.

\paragraph{\textbf{Designing efficient NN model architectures}}
One line of work has focused on optimizing
the NN model architecture in terms of its micro-architecture~\cite{ioannou2017deep,howard2017mobilenets,mamalet2012simplifying,ma2018shufflenet,wu2018shift,sainath2013low,kanjilal1993reduced,zhao2019learning} (e.g., kernel types such as depth-wise convolution or low-rank factorization)
as well as its macro-architecture~\cite{iandola2016squeezenet,howard2017mobilenets,huang2017densely,sandler2018mobilenetv2,howard2019searching,tan2019efficientnet} (e.g., module types such as residual, or inception).
The classical techniques here mostly found new architecture modules
using manual search, which is not scalable. As such, a new line of work is
to design Automated machine learning (AutoML) and Neural Architecture Search (NAS) methods.
These aim to find in an automated way the right NN architecture, under given constraints of model size, depth, and/or width~\cite{zoph2016neural,pham2018efficient,tan2019mnasnet,liu2018darts,wu2019fbnet,wang2020attentivenas}.
We refer interested reader to~\cite{elsken2019neural} for a recent survey of NAS methods.

\paragraph{\textbf{Co-designing NN architecture and hardware together}}
Another recent line of work has been to adapt (and co-design) the NN architecture for
a particular target hardware platform. 
The importance of this is because the overhead of a NN component (in terms of latency and energy) is 
hardware-dependent. 
For example, hardware with a dedicated cache hierarchy can execute bandwidth
bound operations much more efficiently than hardware without such cache hierarchy.
Similar to NN architecture design, initial approaches at architecture-hardware co-design were manual, where an expert would adapt/change the
NN architecture~\cite{gholami2018squeezenext}, followed by using automated
AutoML and/or NAS techniques~\cite{cai2018proxylessnas,cai2019once,wu2019fbnet,howard2019searching}.

\paragraph{\textbf{Pruning}}
Another approach to reducing the memory footprint and computational cost of NNs is to apply pruning.
In pruning, neurons with small \emph{saliency} (sensitivity) are removed, resulting in a sparse computational graph. 
Here, neurons with small saliency are those whose removal minimally affects the model output/loss function.
Pruning methods can be broadly categorized into unstructured pruning~\cite{lecun1990optimal,hassibi1993second,dong2017learning, lee2018snip, xiao2019autoprune, park2020lookahead}, and structured pruning~\cite{luo2017thinet, he2018amc, yu2018nisp, lin2018accelerating, huang2018data, zhao2019variational,yu2021hessian}.
With unstructured pruning, one removes neurons with with small saliency, wherever they occur. 
With this approach, one can perform aggressive pruning, removing most of the NN parameters, with very little impact on the generalization performance of the model.
However, this approach leads to sparse matrix operations, which are known to be hard to accelerate, and which are typically memory-bound~\cite{buluc2008challenges,gale2019state}.
On the other hand, with structured pruning, a group of parameters (e.g., entire convolutional filters) is removed. 
This has the effect of changing the input and output shapes of layers and weight matrices, thus still permitting dense matrix operations. 
However, aggressive structured pruning often leads to significant accuracy degradation.
Training and inference with high levels of pruning/sparsity, while maintaining state-of-the-art performance, has remained an open problem~\cite{blalock2020state}.
We refer the interested reader to~\cite{gale2019state,hoefler2021sparsity,kuzmin2019taxonomy} for a thorough survey of related work in pruning/sparsity.

\paragraph{\textbf{Knowledge distillation}}
Model distillation~\cite{romero2014fitnets, hinton2015distilling, mishra2017apprentice, li2017learning, yim2017gift, polino2018model, ahn2019variational, yin2020dreaming} involves training a large model and then using it as a teacher to train a more compact model. 
Instead of using ``hard'' class labels during the training of the student model, the key idea of model distillation is to leverage the ``soft'' probabilities produced by the teacher, as these probabilities can contain more information about the input. 
Despite the large body of work on distillation, a major challenge here is to achieve a high compression ratio with distillation alone. 
Compared to quantization and pruning, which can maintain the performance with $\geq 4\times$ compression (with INT8 and lower precision), knowledge distillation methods tend to have non-negligible accuracy degradation with aggressive compression. 
However, the combination of knowledge distillation with prior methods (i.e., quantization and pruning) has shown great success~\cite{polino2018model}.

\paragraph{\textbf{Quantization}}
Finally, quantization is an approach that has shown great and consistent success in both training and inference of NN models.
While the problems of numerical representation and quantization are as old as digital computing, 
Neural Nets offer unique opportunities for improvement. 
While this survey on quantization is mostly focused on inference, we should emphasize that 
an important success of quantization has been in NN training~\cite{banner2018scalable, wang2018training, kim2020position, faghri2020adaptive, chmiel2021neural}. In particular,
the breakthroughs of half-precision and mixed-precision training~\cite{courbariaux2014training,gupta2015deep,ginsburg2017tensor,micikevicius2017mixed}
have been the main drivers that have enabled an order of magnitude  higher throughput in AI accelerators.
However, it has proven very difficult to go below half-precision without
significant tuning, and most of the recent quantization research has focused
on inference.
This quantization for inference is the focus of this article.

\paragraph{\textbf{Quantization and Neuroscience}} 
Loosely related to (and for some a motivation for) NN quantization is work in neuroscience that suggests that the human brain stores information in a discrete/quantized form, rather than in a continuous form~\cite{mcculloch1943logical,vanrullen2003perception,tee2020information}.
A popular rationale for this idea is that information stored in continuous form will inevitably get corrupted by noise (which is always present in the physical environment, including our brains, and which can be induced by thermal, sensory, external, synaptic noise, etc.)~\cite{faisal2008noise,chaudhuri2016computational}.
However, discrete signal representations can be more robust to such low-level noise.
Other reasons, including the higher generalization power of discrete representations~\cite{latimer2015single,varshney2016decision,khaw2017discrete}
and their higher efficiency under limited resources~\cite{varshney2006optimal}, have also been proposed.
We refer the reader to~\cite{sun2012framework} for a thorough review of related work in neuroscience literature.

The goal of this work is to introduce current methods and concepts used in quantization and to discuss the current challenges and opportunities 
in this line of research. 
In doing so, we have tried to discuss most relevant work. 
It is not possible to discuss every work in a field as large as NN quantization in
the page limit of a short survey; and there is no doubt that we have missed some relevant papers.
We apologize in advance both to the readers and the authors of papers that we may have neglected.

In terms of the structure of this survey, we will first provide a brief history of quantization in~\sref{sec:history}, and 
then we will introduce basic concepts underlying quantization in~\sref{sec:basic_concepts}.
These basic concepts are shared with most of the quantization algorithms,
and they are necessary for understanding and deploying existing methods.
Then we discuss more advanced topics in~\sref{sec:advanced_concepts}.
These mostly involve recent state-of-the-art methods, especially for low/mixed-precision quantization.
Then we discuss the implications of quantization in hardware accelerators in~\sref{sec:quantization_hardware}, with a special focus on edge processors. 
Finally, we provide a summary and conclusions in~\sref{sec:conclusions}.

\section{General History of Quantization}
\label{sec:history}

Gray and Neuhoff have written a very nice survey of the history of quantization up to 1998 \cite{gray1998quantization}. 
The article is an excellent one and merits reading in its entirety; however, for the reader's convenience we will
briefly summarize some of the key points here.
Quantization, as a method to map from input values in a large (often continuous) set to output values in a small (often finite) set, has a long history.
Rounding and truncation are typical examples.
Quantization is related to the foundations of the calculus, and related methods can be seen in the early 1800s (as well as much earlier), e.g., in early work on least-squares and related techniques for large-scale (by the standards of the early 1800s) data analysis~\cite{Stigler86}.
An early work on quantization dates back to 1867, where discretization was used to approximate the calculation of integrals~\cite{riemann1867ueber}; and, subsequently, in 1897, when Shappard investigated the impact of rounding errors on the integration result~\cite{sheppard1897calculation}.
More recently, quantization has been important in digital signal processing, as the process of representing a signal in digital form ordinarily involves rounding, as well as in numerical analysis and the implementation of numerical algorithms, where computations on real-valued numbers are implemented with finite-precision arithmetic.

It was not until 1948, around the advent of the digital computer, when Shannon wrote his seminal paper on the mathematical theory of communication~\cite{shannon1948mathematical}, that the effect of quantization and its use in coding theory were formally presented.
In particular, Shannon argued in his lossless coding theory that using the same number of bits is wasteful, when events of interest have a non-uniform probability.
He argued that a more optimal approach would be to vary the number of bits based on the probability of an event, a concept that is now known as \emph{variable-rate quantization}.
Huffman coding in particular is motivated by this~\cite{huffman1952method}.
In subsequent work in 1959~\cite{shannon1959coding}, Shannon introduced distortion-rate functions (which provide a lower bound on the signal distortion after coding) as well as the notion of vector quantization (also briefly discussed in~\sref{subsec:vector_quantization}).
This concept was extended and became practical in~\cite{dunn1965performance, gamal1987using, equitz1989new, rose1990deterministic} for real communication applications.
Other important historical research on quantization in signal processing in that time period includes~\cite{oliver1948philosophy}, which introduced
the Pulse Code Modulation (PCM) concept (a pulsing method proposed to
approximate/represent/encode sampled analog signals),
as well as the classical result of high resolution quantization~\cite{bennett1948spectra}.
We refer the interested reader to~\cite{gray1998quantization} for a detailed discussion of these issues.

Quantization appears in a slightly different way in algorithms that use numerical approximation for problems involving continuous mathematical quantities, an area that also has a long history, but that also received renewed interest with the advent of the digital computer.
In numerical analysis, an important notion was (and still is) that of a \emph{well-posed problem}---roughly, a problem is well-posed if: a solution exists; that solution is unique; and that solution depends continuously on the input data in some reasonable topology.
Such problems are sometimes called \emph{well-conditioned problems}.
It turned out that, even when working with a given well-conditioned problem, certain algorithms that solve that problem ``exactly'' in some idealized sense perform very poorly in the presence of ``noise'' introduced by the peculiarities of roundoff and truncation errors.
These roundoff errors have to do with representing real numbers with only finitely-many bits---a quantization specified, e.g., by the IEEE floating point standard; and truncation errors arise since only a finite number of iterations of an iterative algorithm can actually be performed.
The latter are important even in ``exact arithmetic,'' since most problems of continuous mathematics cannot even in principle be solved by a finite sequence of elementary operations; but the former have to do with quantization.
These issues led to the notion of the \emph{numerical stability} of an algorithm.
Let us view a numerical algorithm as a function $f$ attempting to map the input data $x$ to the ``true'' solution $y$; but due to roundoff and truncation errors, the output of the algorithm is actually some other $y^*$.
In this case, the \emph{forward error} of the algorithm is $\Delta y = y^* - y$; and the \emph{backward error} of the algorithm is the smallest $\Delta x$ such that $f(x + \Delta x) = y^*$.
Thus, the forward error tells us the difference between the exact or true answer and what was output by the algorithm; and the backward error tells us what input data the algorithm we ran actually solved exactly.
The forward error and backward error for an algorithm are related by the condition number of the problem.
We refer the interested reader to~\cite{TrefethenBau97} for a detailed discussion of these issues.

\subsection{Quantization in Neural Nets}
\label{subsec:nn-quantization}

No doubt thousands of papers have been written on these topics, and one might wonder: how is recent work on NN quantization different from these earlier works?
Certainly, many of the recently proposed ``novel algorithms'' have strong connections with (and in some cases are essentially rediscoveries of) past work in the literature.
However, NNs bring unique challenges and opportunities to the problem of quantization. 
First, inference and training of Neural Nets are both computationally intensive. So, the efficient representation of numerical values is particularly important. 
Second, most current Neural Net models are heavily over-parameterized, so there is ample opportunity for reducing bit precision without impacting accuracy. 
However, one very important difference is that NNs are very robust to aggressive quantization and extreme discretization. 
The new degree of freedom here has to do with the number of parameters involved, i.e., that we are working with over-parameterized models.
This has direct implications for whether we are solving well-posed problems, whether we are interested in forward error or backward error, etc.
In the NN applications driving recent developments in quantization, there is not a single well-posed or well-conditioned problem that is being solved.
Instead, one is interested in some sort of forward error metric (based on classification quality, perplexity, etc.), but due to the over-parameterization, there are many very different models that exactly or approximately optimize this metric.
Thus, it is possible to have high error/distance between a quantized model and the original non-quantized model, while still attaining very good generalization performance. This added degree of freedom was not present in many of the classical research, which mostly focused on finding compression methods that would not change the signal too much, or with numerical methods in which there was strong control on the difference between the ``exact'' versus the ``discretized'' computation. 
This observation that has been the main driver for researching \emph{novel} techniques for NN~quantization.
Finally,the layered structure of Neural Net models offers an additional dimension to explore. Different layers in a Neural Net have different impact on the loss function, and this motivates a mixed-precision approach to quantization.

\begin{figure}[!t]
    \centering
    \includegraphics[width=0.24\textwidth]{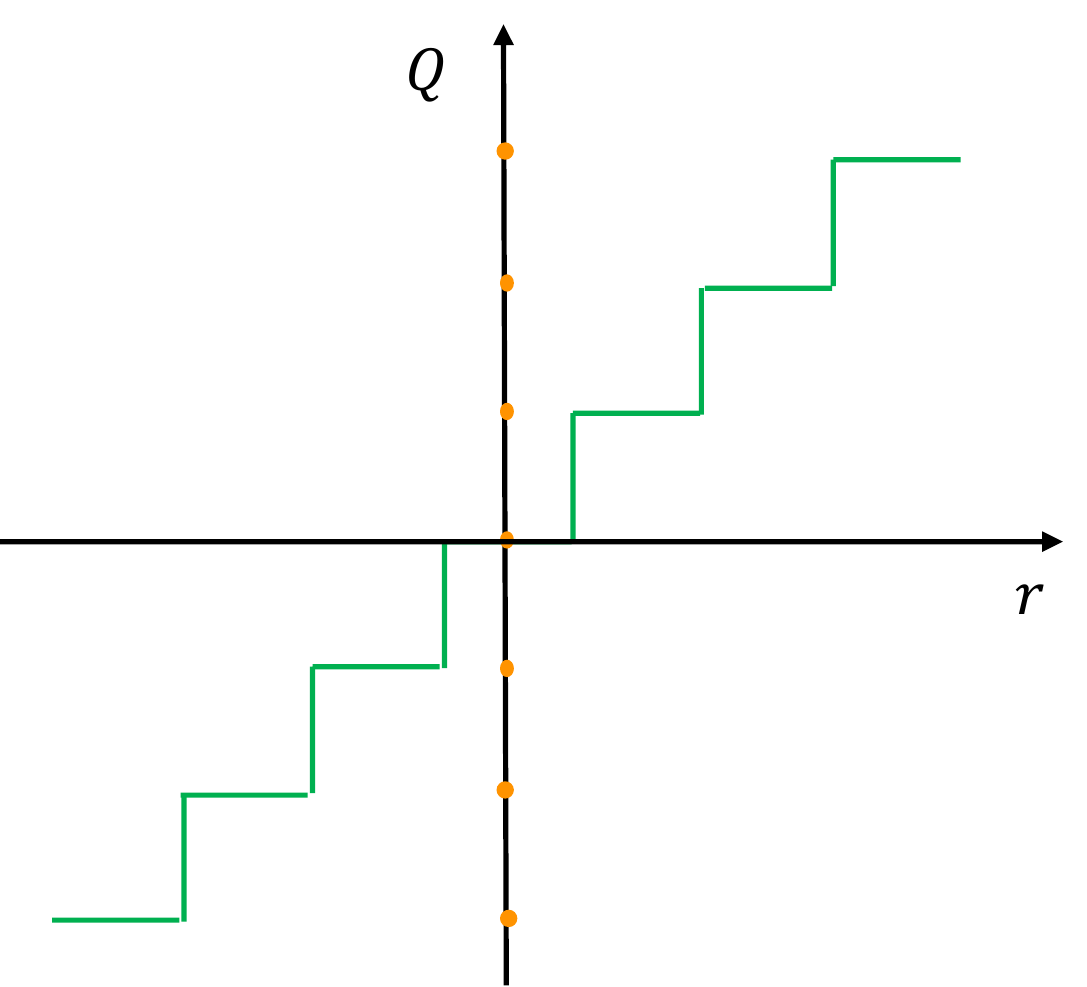}
    \includegraphics[width=0.24\textwidth]{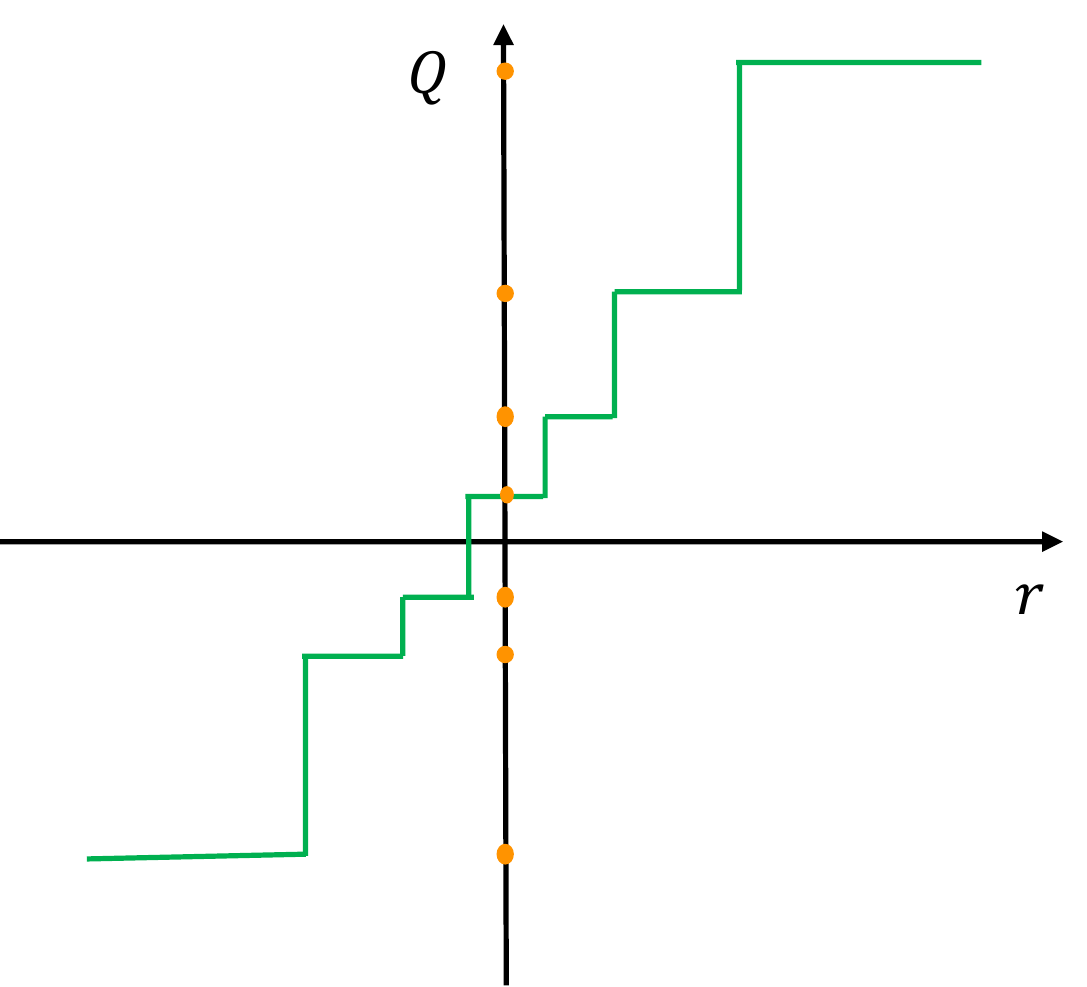}
    \caption{Comparison between uniform quantization (left) and non-uniform quantization (right).
    Real values in the continuous domain $r$ are mapped into discrete, lower precision values in the quantized domain $Q$, which are marked with the orange bullets.
    Note that the distances between the quantized values (quantization levels) are the same in uniform quantization, whereas they can vary in non-uniform quantization.
    }
    \label{fig:uniform}
\end{figure}

\section{Basic Concepts of Quantization}
\label{sec:basic_concepts}
In this section, we first briefly introduce common notations and the problem setup
in~\sref{subsec:problem_setup},
and then we describe the basic quantization concepts and methods in~\sref{subsec:uniform_quant}-\ref{subsec:non_uniform_quant}.
Afterwards, we discuss the different fine-tuning methods in~\sref{subsec:fine_tuning_methods}, followed
by stochastic quantization in~\sref{subsec:stochastic_quantization}.

\subsection{Problem Setup and Notations}
\label{subsec:problem_setup}

Assume that the NN has $L$ layers with learnable parameters, denoted as $\{W_1, W_2, ..., W_L\}$, with $\theta$ denoting the
combination of all such parameters. 
Without loss of generality, we focus on the supervised learning problem, where the nominal goal is to
optimize the following empirical risk minimization function:
\begin{equation}
\small
\label{eq:loss_function}
    \loss(\theta) = \frac{1}{N}\sum_{i=1}^N l(x_i, y_i; \theta),
\end{equation}
where $(x, y)$ is the input data and the corresponding label, $l(x,y;\theta)$ is the loss function
(e.g., Mean Squared Error or Cross Entropy loss),
and $N$ is the total number of data points. 
Let us also denote the input hidden activations of the $i^{th}$ layer as $h_i$,
and the corresponding output hidden activation as $a_i$.
We assume that we have the trained model parameters $\theta$, stored
in floating point precision.
In quantization, the goal is to reduce the precision of both the parameters ($\theta$),
as well as the intermediate activation maps (i.e., $h_i,\ a_i$) to low-precision, 
with minimal impact on the generalization power/accuracy of the model.
To do this, we need to define a quantization operator that maps
a floating point value to a quantized one, which is described next.

\begin{figure*}[!t]
    \centering
    \includegraphics[width=\textwidth]{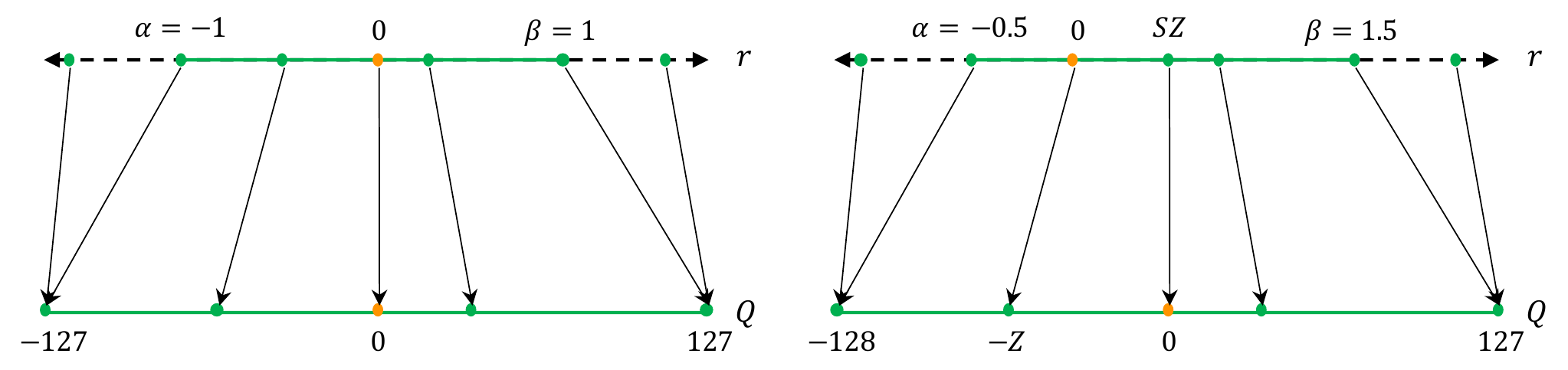}
    \caption{
    Illustration of symmetric quantization and asymmetric quantization.
    Symmetric quantization with restricted range maps real values to [-127, 127], and full range
    maps to [-128, 127] for 8-bit quantization.
    }
    \label{fig:symmetric}
\end{figure*}

\subsection{Uniform Quantization}
\label{subsec:uniform_quant}

We need first to define a function that can quantize
NN weights and activations to a finite set of values.
This function takes real values in floating point, and it maps
them to a lower precision range, as illustrated in~\fref{fig:uniform}.
A popular choice for a quantization function is as follows:
\begin{equation}
\small
\label{eq:quantization_formula}
Q(r) = \text{Int}\big({r}/{S}\big)-Z,
\end{equation}
where $Q$ is the quantization operator, 
$r$ is a real valued input (activation or weight),
$S$ is a real valued scaling factor, 
and $Z$ is an integer zero point.
Furthermore, the $\text{Int}$ function maps a real value to an integer value through a rounding operation (e.g., round to nearest and truncation).
In essence, this function is a mapping from real values $r$ to some integer values.
This method of quantization is also known as \textit{uniform quantization}, as the resulting quantized values (aka quantization levels) are uniformly spaced (\fref{fig:uniform}, left).
There are also \textit{non-uniform quantization} methods whose quantized values are not necessarily uniformly spaced (\fref{fig:uniform}, right), and these methods will be discussed in more detail in \sref{subsec:non_uniform_quant}.
It is possible to recover real values $r$  from the quantized values $Q(r)$ through an operation
that is often referred to as \emph{dequantization}:
\begin{equation}
\small
    \tilde r = S(Q(r)+Z).
\end{equation}
Note that the recovered real values $\tilde{r}$ will not exactly
match $r$ due to the rounding operation.

\begin{figure*}[!t]
    \centering
    \includegraphics[width=0.75\textwidth]{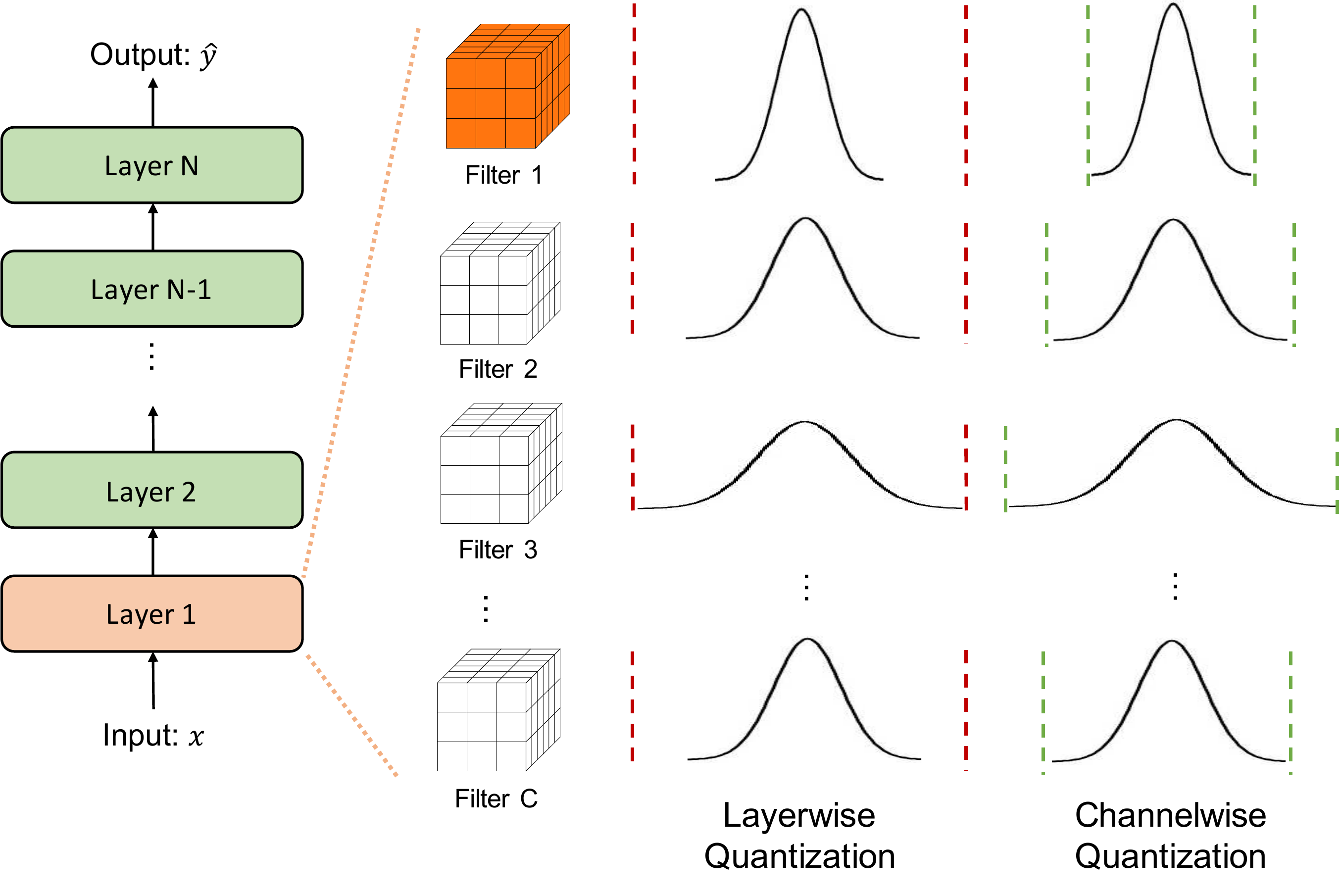}
    \caption{Illustration of different quantization granularities.
    In layerwise quantization, the same clipping range is applied to all the filters that belong to the same layer. 
    This can result in bad quantization resolution for the channels that have narrow distributions (e.g., Filter 1 in the figure).
    One can achieve better quantization resolution using channelwise quantization that dedicates different clipping ranges to different channels.
    }
    \label{fig:quantization_granularity}
\end{figure*}

\subsection{Symmetric and Asymmetric Quantization}
\label{subsec:symmetric_quant}

One important factor in uniform quantization is the choice of the scaling factor $S$ in~\eref{eq:quantization_formula}.
This scaling factor essentially divides a given range of real values $r$ into a number of partitions (as discussed in~\cite{krishnamoorthi2018quantizing, jacob2018quantization}):
\begin{equation}
\small
    S = \frac{\beta - \alpha}{2^{b} - 1},
\end{equation}
where $[\alpha, \beta]$ denotes the clipping range, a bounded range that we are clipping the real values with, and $b$ is the quantization bit width. 
Therefore, in order for the scaling factor to be defined, 
the clipping range $[\alpha, \beta]$ should first be determined. 
The process of choosing the clipping range is often referred to as \textit{calibration}. 
A straightforward choice is to use the min/max of the signal for the clipping range, i.e., $\alpha=r_{min}$, and $\beta=r_{max}$.
This approach is an \emph{asymmetric quantization} scheme, since the clipping range is not necessarily symmetric with respect to the origin, i.e., $-\alpha \ne \beta$, as illustrated in~\fref{fig:symmetric} (Right).
It is also possible to use a \emph{symmetric quantization} scheme by choosing
a symmetric clipping range of $\alpha=-\beta$. 
A popular choice is to choose these
based on the min/max values of the signal:
$-\alpha=\beta={\max(|r_{max}|, |r_{min}|)}$.
Asymmetric quantization often results in a tighter clipping range as compared to symmetric quantization.
This is especially important when the target weights or activations are imbalanced, e.g., the activation after ReLU that always has non-negative values.
Using symmetric quantization, however, simplifies the quantization function in~\eref{eq:quantization_formula} by replacing the zero point with $Z=0$:
\begin{equation}
\small
\label{eq:quantization_formula_our}
Q(r) = \text{Int}\left(\frac{r}{S}\right).
\end{equation}
Here, there are two choices for the scaling factor. In ``full range'' symmetric quantization S is chosen as $\frac{2max(|r|)}{2^n-1}$ (with floor rounding mode),
to use the full INT8 range of [-128,127]. However, in ``restricted range'' S is chosen as $\frac{max(|r|)}{2^{n-1}-1}$, which
only uses the range of [-127,127]. As expected, the full range approach is more accurate.
Symmetric quantization is widely adopted in practice for quantizing weights because zeroing out the zero point can lead to reduction in computational cost during inference~\cite{wu2020integer}, and also makes the implementation more straightforward.
However, note that for activation the cross terms occupying due to the offset in the asymmetric activations are a static data independent term and can be absorbed in the bias (or used to initialize the accumulator)~\cite{bhalgat2020lsq+}.

Using the min/max of the signal for both symmetric and asymmetric quantization is a popular method.
However, this approach is susceptible to outlier data in the activations. 
These could unnecessarily increase the range and, as a result, reduce the resolution of quantization.
One approach to address this is to use percentile instead of min/max of the signal~\cite{mckinstry2018discovering}.
That is to say, instead of the largest/smallest value, the i-th largest/smallest values are used as $\beta$/$\alpha$. 
Another approach is to select $\alpha$ and $\beta$ to minimize KL divergence (i.e., information loss) between the real values and the quantized values~\cite{nvidia8bit}.
We refer the interested readers to~\cite{wu2020integer} where the different calibration methods are evaluated on various~models.

\textbf{Summary (Symmetric vs Asymmetric Quantization).} Symmetric quantization partitions the clipping
using a symmetric range. This has the advantage of easier
implementation, as it leads to $Z=0$ in~\eref{eq:quantization_formula}.
However, it is sub-optimal for cases where the range
could be skewed and not symmetric. For such cases, asymmetric
quantization is preferred.

\subsection{Range Calibration Algorithms: Static vs Dynamic Quantization}
\label{subsec:range_calibration_algorithms}
So far, we discussed different calibration methods for determining the clipping range of $[\alpha, \beta]$.
Another important differentiator of quantization methods is \textit{when} the clipping range is determined. 
This range can be computed statically for weights, as in most cases the parameters are fixed during inference. 
However, the activation maps differ for each input sample ($x$ in~\eref{eq:loss_function}).
As such, there are two approaches to quantizing activations: \textit{dynamic quantization}, and \textit{static quantization}.

In dynamic quantization, this range is \emph{dynamically} calculated for each activation map during runtime.
This approach requires real-time computation of the signal statistics (min, max, percentile, etc.) which
can have a very high overhead. However, dynamic quantization often results in higher
accuracy as the signal range is exactly calculated for each input.

Another quantization approach is static quantization, in which the clipping range is pre-calculated and \emph{static} during inference.
This approach does not add any computational overhead, but it typically results in lower accuracy
as compared to dynamic quantization.
One popular method for the pre-calculation is to run a series of calibration inputs to compute
the typical range of activations~\cite{jacob2018quantization,yao2020hawqv3}.
Multiple different metrics have been proposed to find the best range, including minimizing Mean Squared Error (MSE) between
original unquantized weight distribution and the corresponding quantized values~\cite{sung2015resiliency,shin2016fixed,choukroun2019low,zhao2019improving}.
One could also consider using other metrics such as entropy~\cite{park2017weighted}, although MSE is the most common method used.
Another approach is to learn/impose this clipping range during NN training~\cite{li2019fully, choi2018pact, zhu2016trained, zhang2018lq}.
Notable work here are LQNets~\cite{zhang2018lq}, PACT~\cite{choi2018pact}, LSQ~\cite{esser2019learned}, and LSQ+~\cite{bhalgat2020lsq+}
which jointly optimizes the clipping range and the weights in NN during training.

\textbf{Summary (Dynamic vs Static Quantization).}
Dynamic quantization dynamically computes the clipping range of each
activation and often achieves the highest accuracy. However,
calculating the range of a signal dynamically is very expensive, and
as such, practitioners most often use static quantization where the clipping
range is fixed for all inputs.

\subsection{Quantization Granularity}
\label{subsec:quant_granularity}
In most computer vision tasks, the activation input to a layer is convolved with many different convolutional filters, as illustrated in~\fref{fig:quantization_granularity}.
Each of these convolutional filters can have a different range of values.
As such, one differentiator for quantization methods is the granularity of how the clipping range $[\alpha, \beta]$ is calculated
for the weights. 
We categorized them as follows.

\paragraph{{Layerwise Quantization}} In this approach, the clipping range
is determined by considering all of the weights in convolutional filters of a layer~\cite{krishnamoorthi2018quantizing}, as shown in the third column of~\fref{fig:quantization_granularity}.
Here one examines the statistics of the entire parameters in that layer (e.g., min, max, percentile, etc.),
and then uses the same clipping range for all the convolutional filters.
While this approach is very simple to implement, it
often results in sub-optimal accuracy, as the range of each convolutional filter can
be vary a lot.
For example, a convolutional kernel that has
relatively narrower range of parameters may lose its quantization resolution due to another kernel in the same layer with a wider range.

\paragraph{{Groupwise Quantization}}
One could group multiple different channels inside a layer to calculate the clipping range (of either activations or convolution kernels).
This could be helpful for cases where the distribution of the parameters across a single convolution/activation varies a lot.
For instance, this approach was found useful in Q-BERT~\cite{shen2020q} for quantizing Transformer~\cite{vaswani2017attention} models that consist of fully-connected attention layers.
However, this approach inevitably comes with the extra cost of accounting for
different scaling factors.

\paragraph{{Channelwise Quantization}} 
A popular choice of the clipping range is to use a fixed value for each convolutional filter, independent of
other channels~\cite{zhou2016dorefa, zhang2018lq, jacob2018quantization, krishnamoorthi2018quantizing, huang2021codenet, shkolnik2020robust}, as shown in the last column of~\fref{fig:quantization_granularity}.
That is to say, each channel is assigned a dedicated scaling factor.
This ensures a better quantization resolution and often results in
higher accuracy.

\paragraph{{Sub-channelwise Quantization}}
The previous approach could be taken to the extreme, where the clipping range is determined with respect to any groups of parameters in a convolution or fully-connected layer.
However, this approach could add considerable
overhead, since the different scaling factors need to be taken into account when processing a single convolution or full-connected layer.
Therefore, groupwise quantization could establish a good compromise between the quantization resolution and the computation overhead.

\textbf{Summary (Quantization Granularity).} Channelwise quantization is currently
the standard method used for quantizing convolutional kernels.
It enables the practitioner to adjust the clipping range for
each individual kernel with negligible overhead. In contrast,
sub-channelwise quantization may result in significant overhead
and is not currently the standard choice (we also refer interested
reader to~\cite{garg2021confounding} for tradeoffs
associated with these design choices).

\begin{figure*}[!t]
    \centering
    \includegraphics[width=0.75\textwidth]{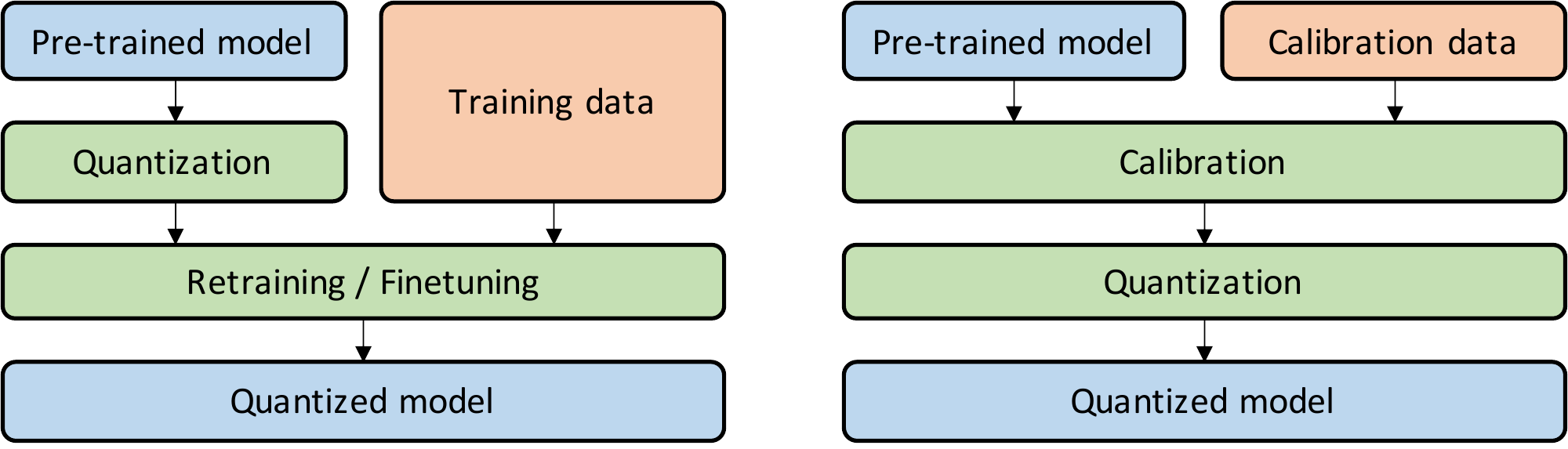}
    \caption{Comparison between Quantization-Aware Training (QAT, Left) and Post-Training Quantization (PTQ, Right). 
    In QAT, a pre-trained model is quantized and then finetuned using training data to adjust parameters and recover accuracy degradation. 
    In PTQ, a pre-trained model is calibrated using calibration data (e.g., a small subset of training data) to compute the clipping ranges and the scaling factors.
    Then, the model is quantized based on the calibration result.
    Note that the calibration process is often conducted in parallel with the finetuning process for QAT.
    }
    \label{fig:quantization_process}
\end{figure*}

\subsection{Non-Uniform Quantization}
\label{subsec:non_uniform_quant}

Some work in the literature has also explored non-uniform
quantization~\cite{gong2014compressing, wu2016quantized, gupta2015deep, hou2016loss, lin2015neural, miyashita2016convolutional, choi2016towards, cai2017deep, park2018value, zhang2018lq, wang2018two, jeon2020biqgemm,jung2019learning,yang2019quantization,faraone2018syq,tung2018clip,zhou2018explicit,park2017weighted, yang2020searching, liao2020sparse},
where quantization steps as well as quantization levels are allowed to be non-uniformly spaced.
The formal definition of non-uniform quantization is shown in~\eref{eq:non_uniform}, where $X_i$ represents the discrete quantization levels and $\Delta_i$ the quantization steps (thresholds): 
\begin{equation}
\small
\label{eq:non_uniform}
    Q(r) = X_i,~\mathrm{if}~r \in [\Delta_i, \Delta_{i+1}).
\end{equation}
Specifically, when the value of a real number $r$ falls in between the quantization step $\Delta_i$ and $\Delta_{i+1}$, quantizer $Q$ projects it to the corresponding quantization level $X_i$.
Note that neither $X_i$'s nor $\Delta_{i}$'s are uniformly spaced.

Non-uniform quantization may achieve higher accuracy for a fixed bit-width, 
because one could better capture the distributions by focusing more on important value regions or 
finding appropriate dynamic ranges. 
For instance, many non-uniform quantization methods have been designed for bell-shaped distributions of the weights and activations that often involve long tails~\cite{baskin2018uniq, cai2017deep, li2019additive, miyashita2016convolutional, jain2019biscaled, fang2020post}. 
A typical rule-based non-uniform quantization is to use a logarithmic distribution~\cite{miyashita2016convolutional, zhou2017incremental}, where the quantization steps and levels increase exponentially instead of linearly. 
Another popular branch is \textit{binary-code-based} quantization~\cite{jeon2020biqgemm, xu2018alternating, zhang2018lq, guo2017network, hubara2016binarized} 
where a real-number vector $\mathbf{r} \in \mathbb{R}^n$ is quantized into $m$ binary vectors by
representing $\mathbf{r} \approx \sum_{i=1}^{m}\alpha_i \mathbf{b}_i$, with the scaling factors $\alpha_i \in \mathbb{R}$ and the binary vectors  $\mathbf{b}_i \in \{-1, +1\}^n$.
Since there is no closed-form solution for minimizing the error between $\mathbf{r}$ and $\sum_{i=1}^{m}\alpha_i \mathbf{b}_i$, 
previous research relies on heuristic solutions.
To further improve the quantizer, more recent work~\cite{tang2017train, guo2017network, xu2018alternating} formulates non-uniform quantization as an optimization problem. As shown in~\eref{eq:optimization_quantization}, the quantization steps/levels in the quantizer $Q$ are adjusted to minimize the difference between the original tensor and the quantized counterpart.
\begin{equation}
\small
\label{eq:optimization_quantization}
    \min_{Q}\|Q(r) - r\|^2
\end{equation}
Furthermore, the quantizer itself can also be jointly trained with the model parameters. These methods are referred to as learnable quantizers, and the quantization steps/levels are generally trained with iterative optimization~\cite{zhang2018lq, xu2018alternating} or gradient descent~\cite{lin2017towards, jung2019learning, yang2019quantization}.

In addition to rule-based and optimization-based non-uniform quantization, clustering can also be beneficial to alleviate the information loss due to quantization.
Some works~\cite{gong2014compressing, wu2016quantized} use k-means on different tensors to determine the quantization steps and levels, while other work~\cite{choi2016towards} applies a Hessian-weighted k-means clustering on weights to minimize the performance loss. Further discussion can be found in~\sref{subsec:vector_quantization}.

\textbf{Summary (Uniform vs Non-uniform Quantization).}
Generally, non-uniform quantization enables us to better capture
the signal information, by assigning bits and discreitizing the range of parameters non-uniformly. 
However, non-uniform quantization schemes are typically difficult to deploy efficiently on general computation hardware, e.g., GPU and CPU.
As such, the uniform quantization is currently the de-facto method due to its
simplicity and its efficient mapping to hardware.

\begin{figure*}[!t]
    \centering
    \includegraphics[width=0.75\textwidth]{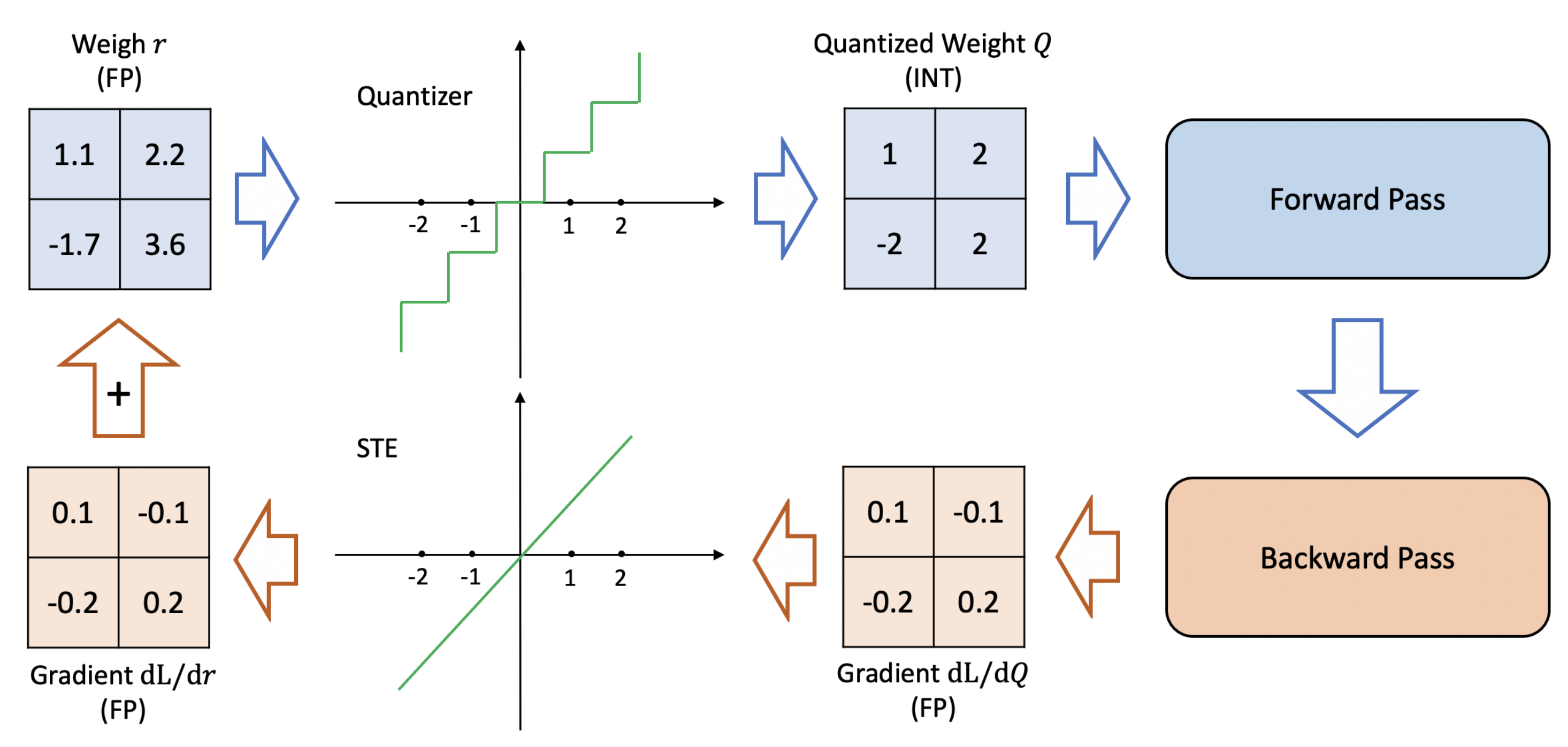}
    \caption{Illustration of Quantization-Aware Training procedure, including
    the use of Straight Through Estimator (STE).
    }
    \label{fig:qat}
\end{figure*}

\subsection{Fine-tuning Methods}
\label{subsec:fine_tuning_methods}
It is often necessary to adjust the parameters in the NN after quantization.
This can either be performed by re-training the model, a process that is called
Quantization-Aware Training (QAT), or done without re-training, a process that is often
referred to as Post-Training Quantization (PTQ). 
A schematic comparison between these two
approaches is illustrated in~\fref{fig:quantization_process}, and further
discussed below (we refer interested reader to~\cite{nagel2021white}
for more detailed discussion on this topic).

\subsubsection{Quantization-Aware Training}
Given a trained model, quantization may introduce a perturbation to the trained model parameters, and this can push the model away from the point to which it had converged when it was trained with floating point precision. 
It is possible
to address this by re-training the NN model with quantized parameters so that the model
can converge to a point with better loss. 
One popular approach is to use Quantization-Aware Training (QAT), 
in which the usual forward and backward pass
are performed on the quantized model in floating point,
but the model parameters are quantized after each gradient
update (similar to projected gradient descent).
In particular, it is important to do this projection
after the weight update is performed in floating point
precision.
Performing the backward pass with floating point is important, 
as accumulating the gradients in quantized precision
can result in zero-gradient or gradients that have high error, especially in low-precision
~\cite{courbariaux2015binaryconnect,lin2015neural,hubara2016binarized,rastegari2016xnor, gysel2016hardware, gysel2018ristretto, tailor2020degree, ni2020wrapnet}. 

An important subtlety in backpropagation is how the the non-differentiable quantization operator (\eref{eq:quantization_formula}) is treated.
Without any approximation, the gradient of this operator is zero almost everywhere, since the rounding operation in~\eref{eq:quantization_formula} is
a piece-wise flat operator.
A popular approach to address this is to approximate the gradient of this operator by the so-called Straight Through Estimator (STE)~\cite{bengio2013estimating}.
STE essentially ignores the rounding operation and approximates it with an identity function, as illustrated in~\fref{fig:qat}.

Despite the coarse approximation of STE, it often works well in practice, except for ultra low-precision quantization
such as binary quantization~\cite{bai2018proxquant}. The work of~\cite{yin2019understanding} provides
a theoretical justification for this phenomena, and it finds that the coarse gradient
approximation of STE can in expectation correlate with population gradient (for a proper choice of STE).
From a historical perspective, we should note that the original
idea of STE can be traced back to the seminal work of~\cite{rosenblatt1957perceptron,rosenblatt1961principles}, where
an identity operator was used to approximate gradient from the binary neurons.

While STE is the mainstream approach~\cite{zhuang2018towards, stock2021training}, other approaches have also been explored in the literature~\cite{chen2019metaquant,fan2020training,cai2017deep,liu2018bi,leng2018extremely, agustsson2020universally}.
We should first mention that~\cite{bengio2013estimating} also proposes a stochastic
neuron approach as an alternative to STE (this is briefly discussed
in~\sref{subsec:stochastic_quantization}).
Other approaches using combinatorial optimization~\cite{friesen2017deep}, target propagation~\cite{lee2015difference}, or 
Gumbel-softmax~\cite{jang2016categorical}
have also been proposed.
Another different class of alternative methods tries to use regularization operators to enforce the weight to be quantized.
This removes
the need to use the non-differentiable quantization operator in~\eref{eq:quantization_formula}.
These
are often 
referred to as \emph{Non-STE} methods~\cite{choi2018learning,bai2018proxquant,naumov2018periodic,leng2018extremely,hou2016loss,zhou2017incremental,alizadeh2020gradient}.
Recent research in this area includes
ProxQuant~\cite{bai2018proxquant} which
removes the rounding operation in the quantization formula~\eref{eq:quantization_formula}, and instead uses
the so-called \emph{W-shape}, non-smooth regularization function to enforce the weights to quantized values.
Other notable research includes using pulse training to approximate the derivative of discontinuous points~\cite{deng2018gxnor},
or replacing the quantized weights with an affine combination of floating point and quantized parameters~\cite{liu2019learning}.
The recent work of~\cite{nagel2020up} also suggests AdaRound, which is an adaptive rounding method as an alternative
to round-to-nearest method.
Despite interesting works in this area, these methods often require a lot of tuning and so far STE approach is the most commonly used method.

In addition to adjusting model parameters,
some prior work found it effective to learn quantization parameters  during QAT as well.
PACT~\cite{choi2018pact} learns the clipping ranges of activations under uniform quantization, while QIT~\cite{jung2019learning} also learns quantization steps and levels as an extension to a non-uniform quantization setting.
LSQ~\cite{esser2019learned} introduces a new gradient estimate to learn scaling factors for non-negative activations (e.g., ReLU) during QAT, and LSQ+~\cite{bhalgat2020lsq+} further extends this idea to general activation functions such as swish~\cite{ramachandran2017searching} and h-swish~\cite{howard2019searching} that produce negative values.

\textbf{Summary (QAT).}
QAT has been shown to work despite the coarse
approximation of STE. However, the main disadvantage of QAT is the computational cost of re-training
the NN model. This re-training may need to be performed for several hundred epochs to recover accuracy,
especially for low-bit precision quantization. 
If a quantized model is going to be deployed for an extended period, and if efficiency and accuracy are especially important,
then this investment in re-training is likely to be worth it. However, this is not always the case,
as some models have a relatively short lifetime. 
Next, we next discuss an alternative approach that does not
have this overhead.

\subsubsection{Post-Training Quantization}
An alternative to the expensive QAT method is Post-Training Quantization (PTQ) which performs the quantization and the adjustments of the weights,
without any fine-tuning~\cite{banner2018post,meller2019same,choukroun2019low,zhao2019improving,fang2020post,fang2020near,lee2018quantization,nagel2019data,cai2020zeroq, li2021brecq, he2018learning, garg2021confounding, garg2021dynamic, hubara2020improving,shomron2021post}.
As such, the overhead of PTQ is very low and often negligible.
Unlike QAT, which requires a sufficient amount of training data for retraining, PTQ has an additional advantage that it can be applied in situations where data is limited or unlabeled.
However, this often comes at the cost of lower accuracy as compared to QAT, especially for low-precision quantization.

For this reason, multiple approaches have been proposed to mitigate the accuracy degradation of PTQ.
For example, \cite{banner2018post, finkelstein2019fighting} observe inherent bias in the mean and variance of the weight values following their quantization and propose bias correction methods; and \cite{meller2019same, nagel2019data} show that equalizing the weight ranges (and implicitly activation ranges)
between different layers or channels can reduce quantization errors. 
ACIQ~\cite{banner2018post} analytically computes the optimal clipping range and the channel-wise bitwidth setting for PTQ. 
Although ACIQ can achieve low accuracy degradation, the channel-wise activation quantization used in ACIQ is hard to efficiently deploy on hardware.
In order to address this, the OMSE method~\cite{choukroun2019low} removes channel-wise quantization on activation and proposes to conduct PTQ by optimizing the L2 distance between the quantized tensor and the corresponding floating point tensor.
Furthermore, to better alleviate the adverse impact of outliers on PTQ, an outlier channel splitting (OCS) method is proposed in~\cite{zhao2019improving} which duplicates and halves the channels containing outlier values.
Another notable work is AdaRound~\cite{nagel2020up} which shows that the naive round-to-nearest method for quantization can counter-intuitively results in sub-optimal solutions, and it proposes an adaptive rounding method that better reduces the loss.
While AdaRound restricts the changes of the quantized weights to be within $\pm1$ from their full-precision counterparts, 
AdaQuant~\cite{hubara2020improving} proposes a more general method that allows the quantized weights to change as needed.
PTQ schemes can be taken to the extreme, where neither training nor testing data are utilized during quantization (aka zero-shot scenarios),
which is discussed next.

\textbf{Summary (PTQ).} In PTQ, all the weights and activations
quantization parameters are determined without any re-training of the
NN model. As such, PTQ is a very fast method for quantizing
NN models. However, this often comes at the cost of lower accuracy
as compared to QAT.

\begin{figure*}[!t]
    \centering
    \includegraphics[width=0.9\textwidth]{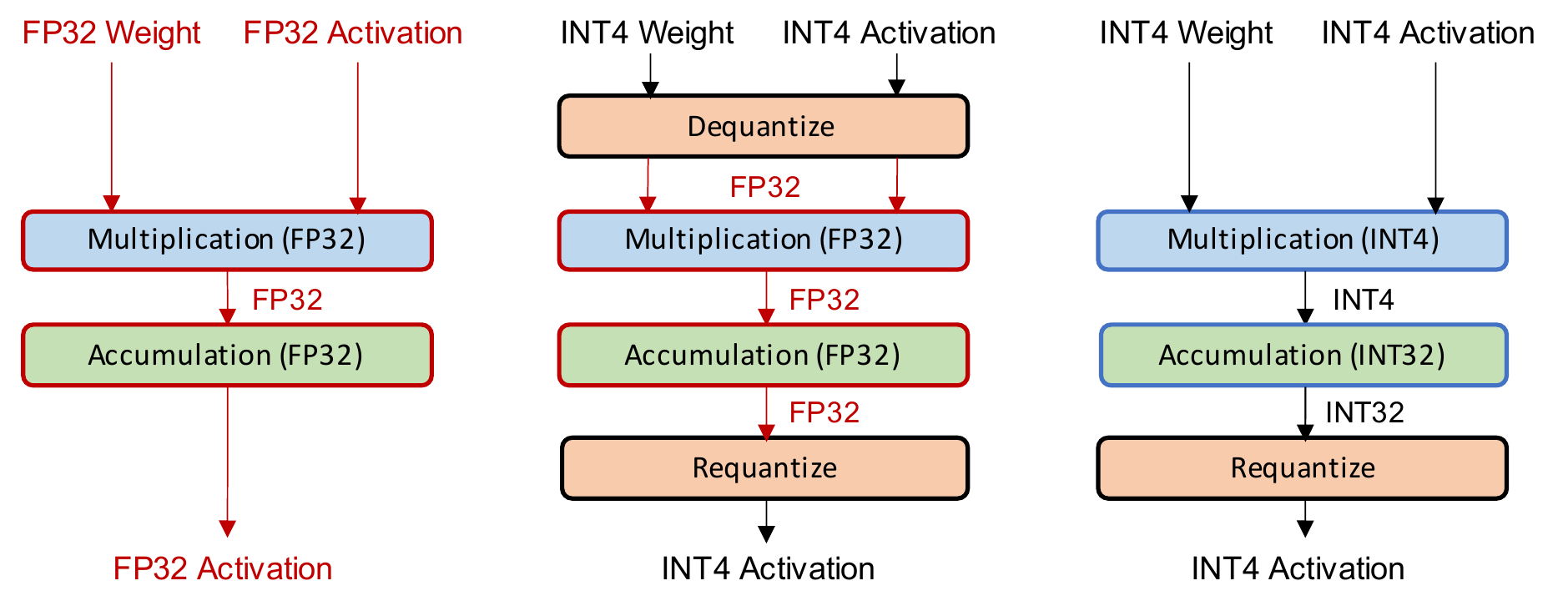}
    \caption{
    Comparison between full-precision inference (Left), inference with simulated quantization (Middle), and inference with integer-only quantization (Right).}
    \label{fig:int_only}
\end{figure*}

\subsubsection{Zero-shot Quantization}
\label{subsubsec:zero_shot_quantization}
As discussed so far, in order to achieve minimal accuracy degradation
after quantization, we need access to the entire of
a fraction of training data.
First, we need to know the range of activations so that we can clip the values and determine the proper scaling factors (which
is usually referred to as calibration in the literature).
Second, quantized models often require fine-tuning to adjust the model parameters and recover the accuracy degradation.
In many cases, however, access to the original training data is not possible during the quantization procedure.
This is because the training dataset is either too large to be distributed,
proprietary (e.g., Google's JFT-300M),
or sensitive due to security or privacy concerns (e.g., medical data).
Several different methods have been proposed to address this challenge, which we refer to as zero-shot quantization (ZSQ). 
Inspired by ~\cite{nagel2019data}, here we first describe two different levels of zero-shot quantization:
\begin{itemize}
    \item \textbf{Level 1:} No data and no finetuning (ZSQ + PTQ). 
    \item \textbf{Level 2:} No data but requires finetuning (ZSQ + QAT).
\end{itemize}
Level 1 allows faster and easier quantization without any finetuning.
Finetuning is in general time-consuming and often requires additional hyperparamenter search.
However, Level 2 usually results in higher accuracy, as finetuning helps the quantized model to recover the accuracy degradation, particularly in ultra-low bit precision settings~\cite{haroush2020knowledge}.
The work of~\cite{nagel2019data} uses a Level 1 approach that relies on 
equalizing the weight ranges and correcting bias errors to make a given NN model more amenable to quantization without any data or finetuning.
However, as this method is based on the scale-equivariance property of (piece-wise) linear activation functions, 
it can be sub-optimal for NNs with non-linear activations, such as BERT~\cite{devlin2018bert} with GELU~\cite{hendrycks2016gaussian} activation or MobileNetV3~\cite{howard2019searching} with swish activation~\cite{ramachandran2017swish}. 

A popular branch of research in ZSQ is to generate synthetic data similar to the real data from which the target pre-trained model is trained.
The synthetic data is then used for calibrating and/or finetuning the quantized model.
An early work in this area \cite{chen2019data} exploits Generative Adversarial Networks (GANs)~\cite{goodfellow2014generative} for synthetic data generation.
Using the pre-trained model as a discriminator, it trains the generator so that its outputs can be well classified by the discriminator.
Then, using the synthetic data samples collected from the generator, 
the quantized model can be finetuned with knowledge distillation from the full-precision counterpart (see \sref{subsec:distillation} for more details).
However, this method fails to capture the internal statistics (e.g., distributions of the intermediate layer activations) of the real data, as it is generated only using the final outputs of the model.
Synthetic data which does not take the internal statistics into account may not properly represent the real data distribution~\cite{haroush2020knowledge}.
To address this, a number of subsequent efforts use the statistics stored in Batch Normalization (BatchNorm)~\cite{ioffe2015batch}, i.e., channel-wise mean and variance, to generate more realistic synthetic data.
In particular, \cite{haroush2020knowledge} generates data by directly minimizing the KL divergence of the internal statistics, and it uses the synthetic data to calibrate and finetune the quantized models.
Furthermore, ZeroQ~\cite{cai2020zeroq} shows that the synthetic data can be used for sensitivity measurement as well as calibration, thereby enabling mixed-precision post-training quantization without any access to the training/validation data.
ZeroQ also extends ZSQ to the object detection tasks, as it does not rely on the output labels when generating data. 
Both~\cite{haroush2020knowledge} and~\cite{cai2020zeroq} set the input images as trainable parameters and directly perform backpropagation on them until their internal statistics become similar to those of the real data.
To take a step further, recent research~\cite{choi2020data, xu2020generative, he2021generative} finds it effective to train and exploit generative models that can better capture the real data distribution and generate more realistic synthetic data.

\textbf{Summary (ZSQ).} Zero Shot (aka data free) quantization
performs the entire quantization without any access to the training/validation
data. This is particularly important for Machine Learning as a Service (MLaaS) providers who want to accelerate the deployment of a customer's workload,
without the need to access their dataset. Moreover, this is important
for cases where security or privacy concerns may limit access to the training
data.
\begin{figure*}[!t]
    \centering
    \includegraphics[width=0.41\textwidth]{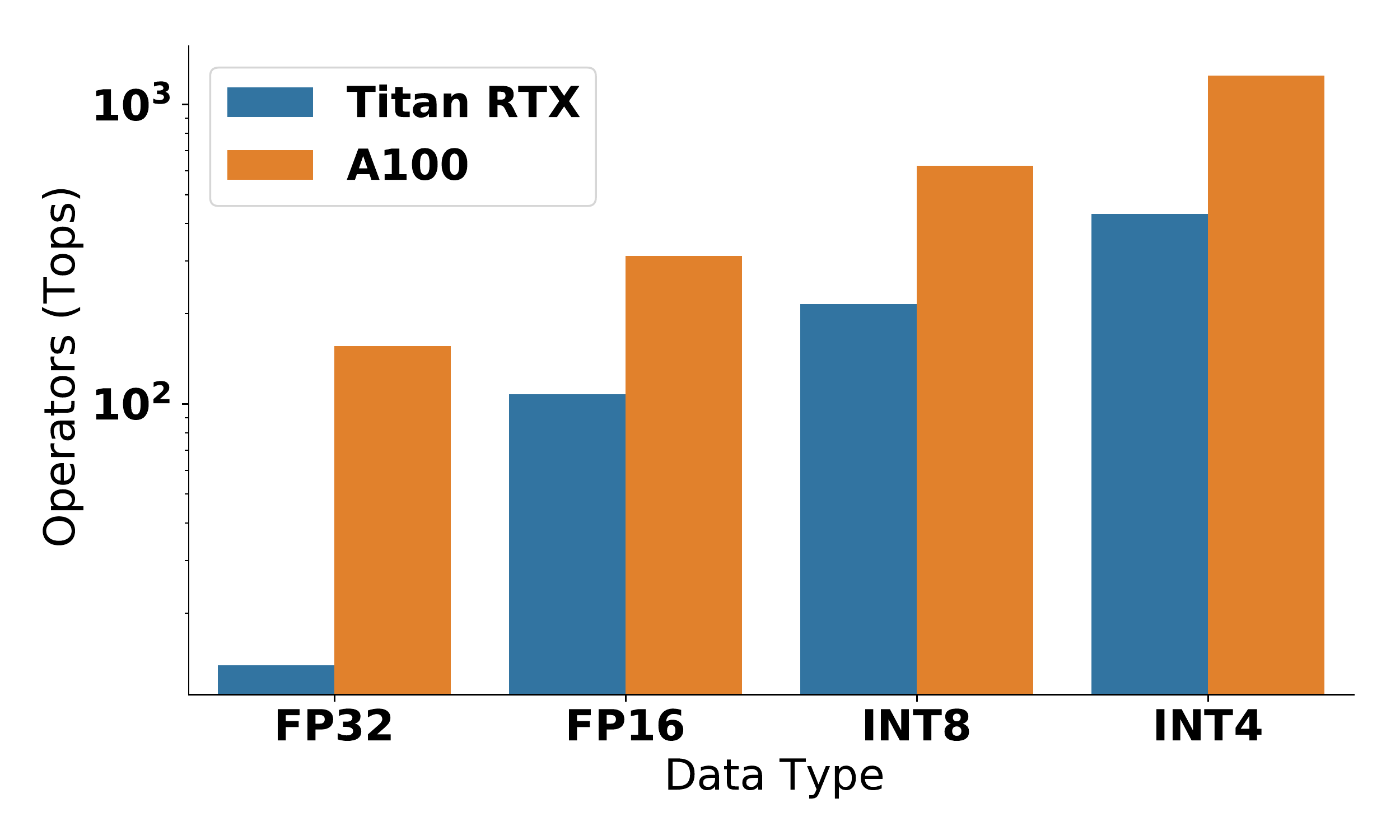}
    \includegraphics[trim=0 60 50 10, clip, width=0.58\textwidth]{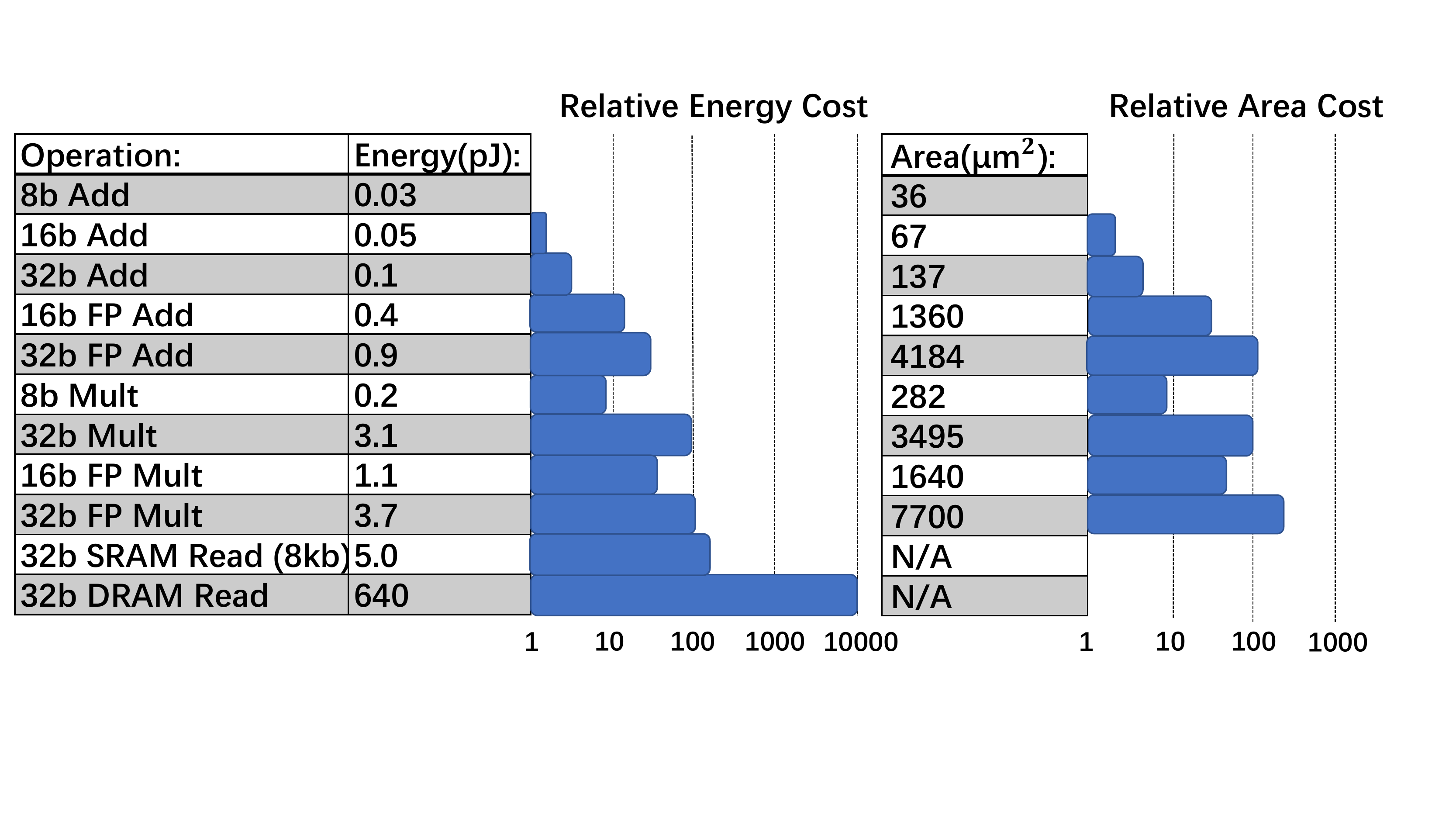}
    \caption{(Left) Comparison between peak throughput for different bit-precision logic on Titan RTX and A100 GPU. 
    (Right) Comparison of the corresponding energy cost and relative area cost for different precision for 45nm technology~\cite{horowitz20141}.
    As one can see, lower precision provides exponentially better energy efficiency and higher throughput.
    }
    \label{fig:gpu_comparison}
\end{figure*}
\subsection{Stochastic Quantization}
\label{subsec:stochastic_quantization}
During inference, the quantization scheme is usually deterministic.
However, this is not the only possibility, and some works have explored stochastic quantization for quantization aware training as well
as reduced precision training~\cite{bengio2013estimating,gupta2015deep}.
The high level intuition has been that the stochastic quantization may
allow a NN to explore more, as compared to deterministic quantization.
One popular supporting argument has been that small weight updates may not lead to any
weight change, as the rounding operation may always return the same weights. However, enabling
a stochastic rounding may provide the NN an opportunity to \emph{escape}, thereby updating
its parameters.

More formally, stochastic quantization maps the floating number up or down with a probability associated
to the magnitude of the weight update.
For instance, in~\cite{gupta2015deep, chen2020statistical}, the $\text{Int}$ operator in~\eref{eq:quantization_formula} is defined as 
\begin{equation}
\small
    \text{Int}(x) = 
    \begin{cases}
    \lfloor x \rfloor \quad \text{with probability } \lceil x \rceil-x,\\ 
    \lceil x \rceil \quad \text{with probability } x - \lfloor x \rfloor.
    \end{cases}
\end{equation}
However, this definition cannot be used for binary quantization. 
Hence,~\cite{courbariaux2015binaryconnect} extends this to 
\begin{equation}
\small
    \text{Binary}(x) = 
    \begin{cases}
    -1 \quad \text{with probability } 1-\sigma(x),\\ 
    +1 \quad \text{with probability } \sigma(x),
    \end{cases}
\end{equation}
where $\text{Binary}$ is a function to binarize the real value $x$, and $\sigma(\cdot)$ is the sigmoid function. 

Recently, another stochastic quantization method is introduced in QuantNoise~\cite{fan2020training}. 
QuantNoise quantizes a different random subset of weights during each forward pass and trains the model with unbiased gradients.
This allows lower-bit precision quantization without significant accuracy drop in many computer vision and natural language processing models.
However, a major challenge with stochastic quantization methods is the overhead of creating
random numbers for every single weight update, and as such they are not yet adopted widely in practice.

\begin{figure*}[!t]
    \centering
    \includegraphics[width=0.9\textwidth]{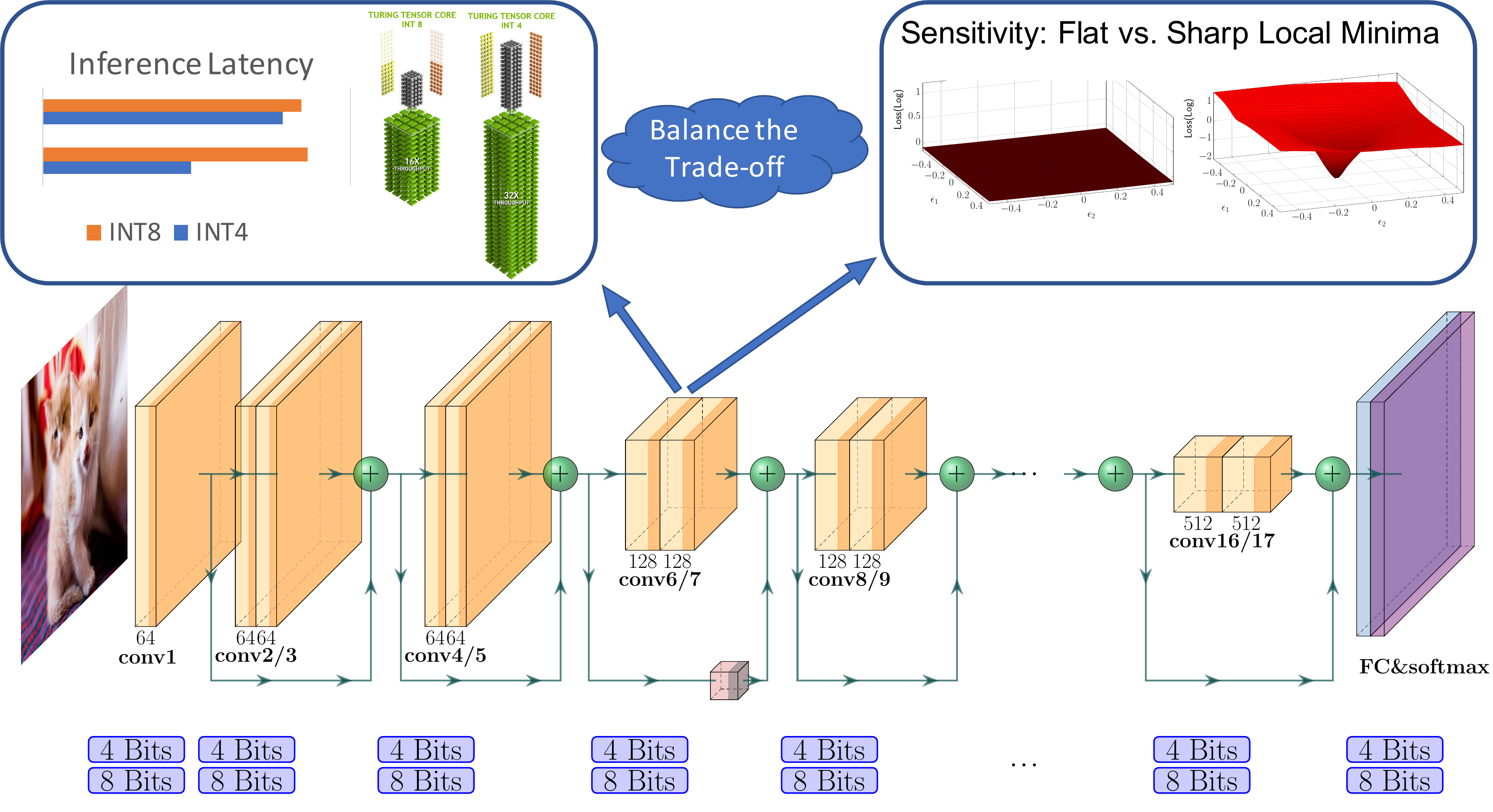}
    \caption{Illustration of mixed-precision quantization. In mixed-precision quantization
    the goal is to keep sensitive and efficient layers in higher precision,
    and only
    apply low-precision quantization to insensitive and inefficient layers.
    The efficiency metric is hardware dependant, and it could be latency or
    energy consumption.
    }
    \label{fig:mixed_precision}
\end{figure*}

\section{Advanced Concepts: Quantization Below 8 bits}
\label{sec:advanced_concepts}

In this section, we will discuss more advanced topics in quantization which are
mostly used for sub-INT8 quantization. We will first discuss
simulated quantization and its difference with integer-only quantization in~\sref{subsec:simulated_and_integer_quantization}.
Afterward, we will discuss different methods for mixed-precision
quantization in~\sref{subsec:mixed_precision_quantization}, followed
by hardware-aware quantization in~\sref{subsec:hardware_aware_quantization}.
Then we will describe how distillation can be used to boost the quantization
accuracy in~\sref{subsec:distillation}, and then we will discuss extremely
low bit precision quantization in~\sref{subsec:extreme_quantization}.
Finally, we will briefly describe the different methods for
vector quantization in~\sref{subsec:vector_quantization}.

\subsection{Simulated and Integer-only Quantization}
\label{subsec:simulated_and_integer_quantization}

There are two common approaches to deploy a quantized NN model, \textit{simulated quantization} (aka fake quantization) and
\textit{integer-only quantization} (aka fixed-point quantization).
In simulated quantization, the quantized model parameters are stored in low-precision, but the operations
(e.g. matrix multiplications and convolutions) are carried out with floating point arithmetic. 
Therefore, the quantized parameters need to be dequantized before the floating point operations as schematically shown in~\fref{fig:int_only} (Middle).
As such, one cannot fully benefit from fast and efficient low-precision logic with simulated quantization.
However, in integer-only quantization, all the operations are performed using low-precision integer arithmetic~\cite{jacob2018quantization,yao2020hawqv3,kim2021bert,lin2016fixed,peng2021fully}, as illustrated in~\fref{fig:int_only} (Right).
This permits the entire inference to be carried out with efficient integer arithmetic, without any floating point dequantization of any parameters or activations.

In general, performing the inference in full-precision with floating point arithmetic may help the final quantization accuracy, but this comes at the cost of not being
able to benefit from the low-precision logic.
Low-precision logic has multiple benefits over the full-precision counterpart in terms of latency, power consumption, and area efficiency.
As shown in~\fref{fig:gpu_comparison} (left), many hardware processors, including NVIDIA V100 and Titan RTX, support fast processing of low-precision arithmetic that can boost the inference throughput and latency.
Moreover, as illustrated in~\fref{fig:gpu_comparison} (right) for a 45nm technology~\cite{horowitz20141}, low-precision logic is significantly
more efficient in terms of energy and area. 
For example, performing INT8 addition is $30\times$ more energy efficient and
$116\times$ more area efficient as compared to FP32 addition~\cite{horowitz20141}.

Notable integer-only quantization works include~\cite{lin2016fixed}, which fuses Batch Normalization into the 
previous convolution layer, and~\cite{jacob2018quantization}, which proposes an integer-only computation method for residual networks with batch normalization. However, both methods are limited to ReLU activation.
The recent work of~\cite{kim2021bert} addresses this limitation by approximating GELU~\cite{hendrycks2016gaussian}, Softmax, and Layer Normalization~\cite{ba2016layer} with integer arithmetic and further extends integer-only quantization to Transformer~\cite{vaswani2017attention} architectures.

\textit{Dyadic quantization} is another class of integer-only quantization,
where all the scaling is performed with dyadic numbers, which
are rational numbers with integer values in their numerator and a power of 2 in the denominator~\cite{yao2020hawqv3}.
This results in a computational graph that only requires integer addition, multiplication, bit shifting, but no integer division.
Importantly, in this approach, all the additions (e.g. residual connections) are enforced to have the same dyadic scale, which
can make 
the addition logic simpler with higher~efficiency.

\textbf{Summary (Simulated vs Integer-only Quantization).}
In general integer-only and dyadic quantization are more desirable as compared
to simulated/fake quantization. 
This is because integer-only uses lower precision logic for the arithmetic,
whereas simulated quantization uses floating point logic to perform
the operations.
However, this does not mean that fake quantization
is never useful. In fact, fake quantization methods can be beneficial for problems
that are bandwidth-bound rather than compute-bound, such as in recommendation systems~\cite{naumov2019deep}.
For these tasks, the bottleneck is the memory footprint and the cost of loading parameters from memory. 
Therefore, performing fake quantization can be acceptable for these cases.

\subsection{Mixed-Precision Quantization}
\label{subsec:mixed_precision_quantization}

It is easy to see that the hardware performance improves as we use lower precision quantization.
However, uniformly quantizing a model to ultra low-precision can cause significant accuracy degradation. 
It is possible to address this with mixed-precision quantization~\cite{zhou2017adaptive,wang2018haq,dong2019hawq,van2020bayesian, yang2021bsq,Qu_2020_CVPR, wang2020apq, hu2021opq, ning2021simple, habi2020hmq, rusci2020leveraging,liu2021layer,zhao2021distribution}. 
In this approach, each layer is quantized with different bit precision, as illustrated in~\fref{fig:mixed_precision}.
One challenge with this approach is that the search space for choosing this bit setting is exponential in the number of layers. 
Different approaches have been proposed to address this huge search space. 

Selecting this mixed-precision for each layer is essentially a searching problem, and many
different methods have been proposed for it.
The recent work of~\cite{wang2018haq} proposed a reinforcement learning (RL) based method to determine automatically the quantization policy, and the authors used a hardware simulator to take the hardware accelerator’s feedback in the RL agent feedback. 
The paper \cite{wu2018mixed} formulated the mixed-precision configuration searching problem as a Neural Architecture Search (NAS) problem and used the Differentiable NAS (DNAS) method to efficiently explore the search space.
One disadvantage of these exploration-based methods~\cite{wang2018haq,wu2018mixed} is that they often
require large computational  resources, and their performance is typically sensitive to hyperparameters and even initialization. 

Another class of mixed-precision methods uses periodic function regularization to train mixed-precision models by automatically 
distinguishing different layers and their varying importance with respect to accuracy while learning their respective bitwidths~\cite{naumov2018periodic}. 

Different than these  exploration and regularization-based approaches, HAWQ~\cite{dong2019hawq} introduces an automatic way to find
the mixed-precision settings based on second-order sensitivity of the model.
It was theoretically shown that the trace of the second-order operator (i.e., the Hessian)
can be used to measure the sensitivity of a layer to quantization~\cite{dong2019hawqv2}, similar to results for
pruning in the seminal work of Optimal Brain Damage~\cite{lecun1990optimal}.
In HAWQv2, this method was extended to mixed-precision activation quantization~\cite{dong2019hawqv2}, 
and was shown to be more than 100x faster than RL based mixed-precision methods~\cite{wang2018haq}.
Recently, in HAWQv3, an integer-only, hardware-aware quantization was introduced~\cite{yao2020hawqv3}
that proposed a fast Integer Linear Programming method to find the optimal
bit precision for a given application-specific constraint (e.g., model size or latency).
This work also addressed the common question about hardware efficiency of mixed-precision
quantization by directly deploying them on T4 GPUs, showing up to 50\% speed up with mixed-precision (INT4/INT8) quantization as compared to
INT8 quantization.

\textbf{Summary (Mixed-precision Quantization).} Mixed-precision quantization has
proved to be an effective and hardware-efficient method for low-precision quantization
of different NN models. In this approach, the layers of a NN are grouped
into sensitive/insensitive to quantization, and higher/lower bits
are used for each layer. As such, one can minimize accuracy degradation
and still benefit from reduced memory footprint and faster speed up
with low precision quantization. Recent work~\cite{yao2020hawqv3} has also shown that this
approach is hardware-efficient as mixed-precision is only used
across operations/layers.

\subsection{Hardware Aware Quantization}
\label{subsec:hardware_aware_quantization}

One of the goals of quantization is to improve the inference latency.
However, not all hardware provide the same speed up after a certain layer/operation is quantized.
In fact, the benefits from quantization is hardware-dependant, with many factors such
as on-chip memory, bandwidth, and cache hierarchy affecting the quantization speed up.

It is important to consider this fact for achieving optimal benefits through
hardware-aware quantization~\cite{wu2016quantized,wang2018haq,he2018amc,wu2018mixed, yang2018netadapt,  yao2020hawqv3, wang2020differentiable, hawks2021ps}.
In particular, the work~\cite{wang2018haq} uses a reinforcement learning agent to determine the hardware-aware mixed-precision setting for quantization, based on a look-up table of latency with respect to different layers with different bitwidth.
However, this approach uses simulated hardware latency. 
To address this the recent work of~\cite{yao2020hawqv3} directly deploys quantized operations
in hardware, and measures the actual deployment latency of each layer for different quantization
bit precisions.

\subsection{Distillation-Assisted Quantization}
\label{subsec:distillation}

An interesting line of work in quantization is to incorporate model distillation to boost quantization accuracy~\cite{polino2018model, mishra2017apprentice, yao2020hawqv3, kadambicomparing}.
Model distillation~\cite{romero2014fitnets, hinton2015distilling, mishra2017apprentice, li2017learning, polino2018model, ahn2019variational, yin2020dreaming, ye2020distillation, zhuang2018towards} is
a method in which a large model with higher accuracy is used as a teacher to
help the training of a compact student model. 
During the training of the student model, instead of using just the ground-truth class labels, model distillation proposes to leverage the soft probabilities produced by the teacher, 
which may contain more information of the input. That is the overall loss function incorporates both the student loss and the distillation loss, which is typically formulated as follows:
\begin{equation}
\label{eq:distillation}
\small
    \mathcal{L} = \alpha \mathcal{H}(y, \sigma(z_s)) + \beta \mathcal{H}(\sigma(z_t, T), \sigma(z_s, T))
\end{equation}

In~\eref{eq:distillation}, $\alpha$ and $\beta$ are weighting
coefficients to tune the amount of loss from the student model and the distillation loss, $y$ is the ground-truth class label, $\mathcal{H}$ is the cross-entropy loss function,
$z_s$/$z_t$ are logits generated by the student/teacher model,
$\sigma$ is the softmax function, and T is its temperature defined as follows:

\begin{equation}
\small
    p_i = \frac{\exp{\frac{z_i}{T}}}{\sum_j \exp{\frac{z_j}{T}}}
\end{equation}

Previous methods of knowledge distillation focus on exploring different knowledge sources.~\cite{hinton2015distilling, li2017learning, park2019relational} use logits (the soft probabilities) as the source of knowledge, while~\cite{romero2014fitnets, yim2017gift, ahn2019variational} try to leverage the knowledge from intermediate layers. The choices of teacher models are also well studied, where~\cite{you2017learning, tarvainen2017mean} use multiple teacher models to jointly supervise the student model, while~\cite{crowley2018moonshine, zhang2019your} apply self-distillation without an extra teacher model.

\subsection{Extreme Quantization}
\label{subsec:extreme_quantization}
Binarization, where the quantized values are constrained to a 1-bit representation, thereby drastically reducing the memory requirement by 32$\times$, is the most extreme quantization method. 
Besides the memory advantages, binary (1-bit) and ternary (2-bit) operations can often be computed efficiently with bit-wise arithmetic and can 
achieve significant acceleration over higher precisions, such as FP32 and INT8.
For instance, the peak binary arithmetic on  NVIDIA V100 GPUs is 8x higher than INT8.
However, a naive binarization method would lead to significant accuracy degradation.
As such, there is a large body of work that has proposed different solutions to address this~\cite{kim2016bitwise, kwon2020structured,wang2020apq,yang2020automatic,jin2020adabits,qin2020forward,zhuang2019structured,wang2019learning,liu2019circulant,zhu2019binary,xu2019main,he2019simultaneously,cai2017deep,guo2017network,juefei2017local,duan2017learning, helwegen2019latent, lee2020flexor, shekhovtsov2020path, jia2020efficient, lin2020rotated, kim2020binaryduo, lidms, han2020training, qin2020bipointnet, bulat2020high, diffenderfer2021multiprize,guo2021boolnet,razani2021adaptive}.

An important work here is
BinaryConnect~\cite{courbariaux2015binaryconnect} which constrains the weights to either +1 or -1.
In this approach, the weights are kept as real values and are only binarized during the forward and backward passes to simulate the binarization effect.
During the forward pass, the real-value weights are converted into +1 or -1 based on the sign function.
Then the network can be trained using the standard training method with STE to propagate the gradients through the non-differentiable sign function. Binarized NN~\cite{hubara2016binarized} (BNN) extends this idea by binarizing the activations as well as the weights.
Jointly binarizing weights and activations has the additional benefit of improved latency, 
since the costly floating-point matrix multiplications can be replaced with lightweight XNOR operations followed by bit-counting.
Another interesting work is Binary Weight Network (BWN) and XNOR-Net proposed in~\cite{deng2018gxnor},
which achieve higher accuracy by incorporating a scaling factor to the weights and using +$\alpha$ or -$\alpha$ instead of +1 or -1. 
Here, $\alpha$ is the scaling factor chosen to minimize the distance between the real-valued weights and the resulting binarized weights.
In other words, a real-valued weight matrix $W$ can be formulated as $W \approx \alpha B$, where $B$ is a binary weight matrix that satisfies the following optimization problem:
\begin{equation}
\small
\label{eq:binarization}
       \alpha, B = \mathrm{argmin} \|W - \alpha B\|^2.
\end{equation}

Furthermore, inspired by the observation that many learned weights are close to zero, there have been attempts to ternarize network by constraining the weights/activations with ternary values, e.g., +1, 0 and -1, thereby explicitly permitting the quantized values to be zero~\cite{lin2015neural, li2016ternary}.
Ternarization also drastically reduces the inference latency by eliminating the costly matrix multiplications as binarization does.
Later, Ternary-Binary Network (TBN)~\cite{wan2018tbn} shows that 
combining binary network weights and ternary activations can achieve an optimal tradeoff between the accuracy and 
computational efficiency.

Since the naive binarization and ternarization methods generally result in severe accuracy degradation, especially for complex tasks such as ImageNet classification, 
a number of solutions have been proposed to reduce the accuracy degradation in extreme quantization.
The work of~\cite{qin2020binary} broadly categorizes these solutions into three branches. 
Here, we briefly discuss each branch, and we refer the interested readers to~\cite{qin2020binary} for more details.

\paragraph{{Quantization Error Minimization}}
The first branch of solutions aims to minimize the quantization error, i.e., the gap between the real values and the quantized values~\cite{li2017performance, hu2018hashing, lin2017towards, liu2018bi, mishra2017wrpn, chin2020one, shen2020balanced, bulat2019xnor, martinez2020training, faraone2018syq, wang2018two}.
Instead of using a single binary matrix to represent real-value weights/activations, 
HORQ~\cite{li2017performance} and ABC-Net~\cite{lin2017towards} use a linear combination of multiple binary matrices, i.e., 
$W \approx \alpha_1 B_1 + \cdots + \alpha_M B_M $, to reduce the quantization error.
Inspired by the fact that binarizing the activations 
reduces their representational capability for the succeeding convolution block,
\cite{mishra2017wrpn} and \cite{chin2020one} show that binarization of wider networks (i.e., networks with larger number of filters) can achieve a good trade-off between the accuracy and the model size.

\paragraph{Improved Loss function} 
Another branch of works focuses on the choice of loss function~\cite{hou2016loss, hou2018loss, zhou2018explicit, ding2019regularizing, wang2019learning}. 
Important works here are loss-aware binarization and ternarization \cite{hou2016loss, hou2018loss} that directly minimize the loss with respect to the binarized/ternatized weights.
This is different from other approaches that only approximate the weights and do not consider the final loss.
Knowledge distillation from full-precision teacher models has also been shown as a promising method to recover the accuracy degradation after binarization/ternarization~\cite{chen2018distilled, polino2018model, xu2019main, mishra2017apprentice}.

\paragraph{Improved Training Method} 
Another interesting branch of work aims for better training methods for binary/ternary models~\cite{darabi2018bnn+, liu2019circulant, gong2019differentiable, bulat2019improved, liu2018bi, zhou2016dorefa, zhu2019binary, alizadeh2018empirical}.
A number of efforts point out the limitation of STE in backpropagating gradients through the sign function: 
STE only propagate the gradients for the weights and/or activations that are in the range of [-1, 1].
To address this, BNN+~\cite{darabi2018bnn+} introduces a continuous approximation for the derivative of the sign function, 
while \cite{xu2019accurate, qin2020forward, yin2019blended} replace the sign function with smooth, differentiable functions that gradually sharpens and approaches the sign function.
Bi-Real Net~\cite{liu2018bi} introduces identity shortcuts connecting activations to activations in consecutive blocks, through which 32-bit activations can be propagated.
While most research focuses on reducing the inference time latency, DoReFa-Net~\cite{zhou2016dorefa} quantizes the gradients in addition to the weights and activations, in order to accelerate the training as well.

Extreme quantization has been successful in drastically reducing the inference/training latency as well as the 
model size for many CNN models on computer vision tasks.
Recently, there have been attempts to extend this idea to Natural Language Processing (NLP) tasks~\cite{zhang2020ternarybert, bai2020binarybert, jin2021kdlsq,ji2021distribution}.
Considering the prohibitive model size and inference latency of state-of-the-art NLP models
(e.g., BERT~\cite{devlin2018bert}, RoBERTa~\cite{liu2019roberta}, and the GPT family~\cite{radford2018improving,radford2019language,brown2020language})
that are pre-trained on a large amount of unlabeled data,
extreme quantization is emerging as a powerful tool for bringing NLP inference tasks to the edge.

\textbf{Summary (Extreme Quantization).}
Extreme low-bit precision quantization is a very promising line of research.
However, existing methods often incur high accuracy degradation as compared
to baseline, unless very extensive tuning and hyperparameter search is performed.
But this accuracy degradation may be acceptable for less critical applications.

\subsection{Vector Quantization}
\label{subsec:vector_quantization}

As discussed in~\sref{sec:history}, quantization has not been invented in machine learning, but has
been widely studied in the past century in information theory,
and particularly in digital signal processing field as a compression tool.
However, the main difference between quantization methods for machine learning is that fundamentally
we are not interested to compress the signal with minimum change/error as compared to the original
signal. Instead, the goal is to find a reduced-precision representation that results
in as small loss as possible. As such, it is completely acceptable if the quantized weights/activations
are far away from the non-quantized ones.

Having said that, there are a lot of interesting ideas
in the classical quantization methods in DSP that have been applied to NN quantization, and in particular
vector quantization~\cite{ballard1999introduction}.
In particular, the work of~\cite{jegou2010product, gong2014compressing, wu2016quantized, park2017weighted, han2015deep, agustsson2017soft, martinez2018lsq++, mukherjee2018biresolution, chen2021incremental} clusters the weights into different groups and use the centroid of each group as quantized values during inference. As shown in~\eref{eq:vector_quantization}, $i$ is the index of weights in a tensor, $c_1, ..., c_k$ are the $k$ centroids found by the clustering, and $c_j$ is the corresponding centroid to $w_i$. After clustering, weight $w_i$ will have a cluster index $j$ related to $c_j$ in the codebook (look-up table). 
\begin{equation}
\label{eq:vector_quantization}
\small
    \min_{c_1, ..., c_k} \sum_i \|w_i - c_j\|^2
\end{equation}
It has been found that using a k-means clustering is sufficient to reduce the model size up to $8\times$ without significant accuracy degradation~\cite{gong2014compressing}. 
In addition to that, jointly applying k-means based vector quantization with pruning and Huffman coding can further reduce the model size~\cite{han2015deep}.

Product quantization~\cite{gong2014compressing, wu2016quantized,stock2019and} is an extension of vector quantization, where the weight matrix is divided into submatrices and vector quantization is applied to each submatrix. 
Besides basic product quantization method, more fine-grained usage of clustering can further improve the accuracy. For example, in~\cite{gong2014compressing} the residuals after k-means product quantization are further recursively quantized. 
And in~\cite{park2017weighted}, the authors apply more clusters for more important quantization ranges to better preserve the information.

\section{Quantization and Hardware Processors}
\label{sec:quantization_hardware}

\begin{figure*}[!t]
    \centering
    \includegraphics[width=\textwidth]{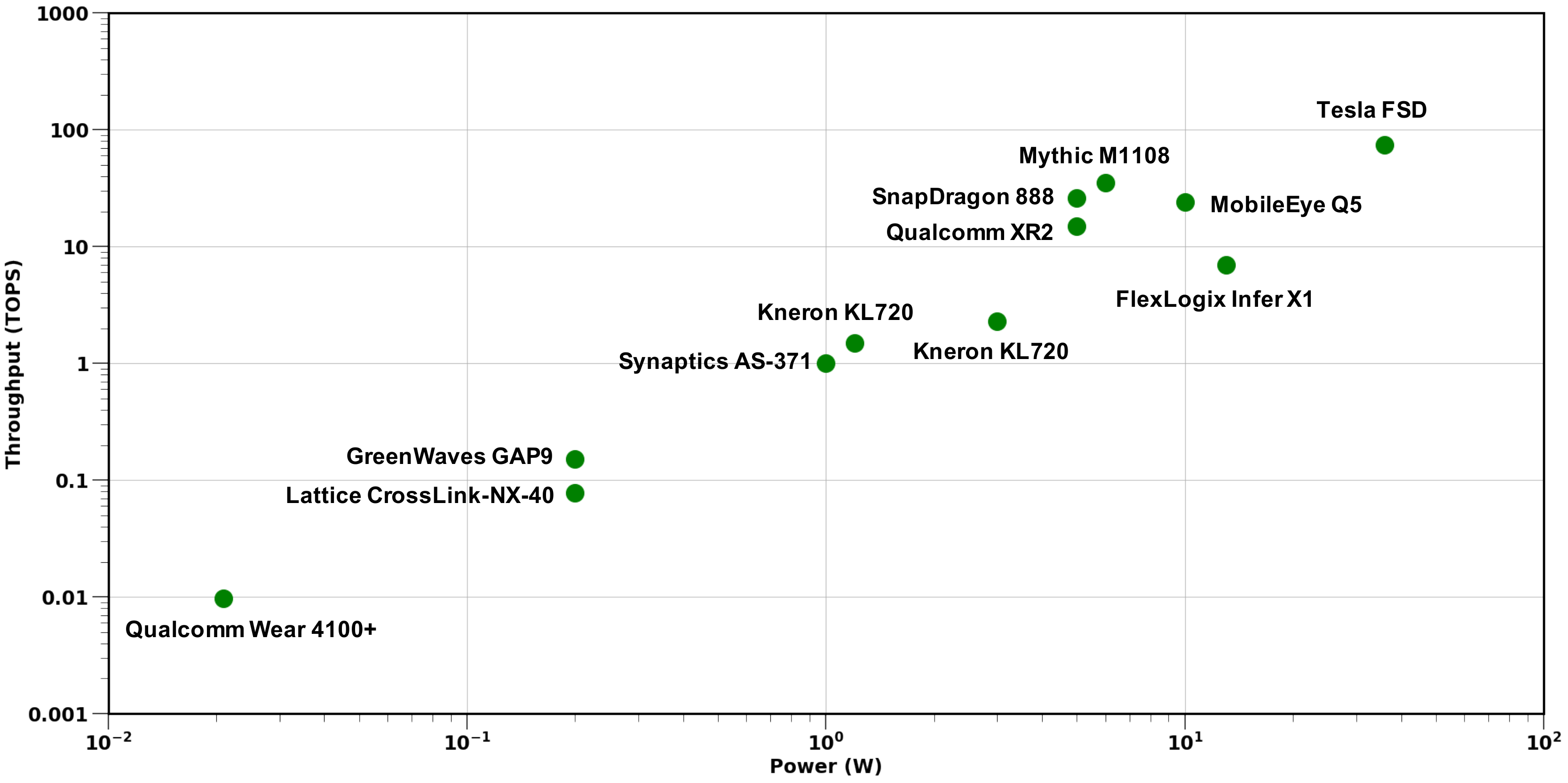}
    \caption{ 
    Throughput comparison of different commercial edge processors for NN inference at the edge.
    }
    \label{fig:hw}
\end{figure*}

We have said that quantization not only reduces the model size, but it also enables faster speed and requires less power, in particular for hardware that has low-precision logic.
As such, quantization has been particularly crucial for edge deployment in IoT and mobile applications.
Edge devices often have tight resource constraints including compute, memory, and importantly power budget.
These are often too costly to meet for many deep NN models.
In addition, many edge processors do not have any support floating point operations, especially in micro-controllers.

Here, we briefly discuss different hardware
platforms in the context of quantization.
ARM Cortex-M is a group of 32-bit RISC ARM processor cores that are designed for low-cost and power-efficient embedded devices. 
For instance, the STM32 family are the microcontrollers based on the ARM Cortex-M cores that are also used for NN inference at the edge.
Because some of the ARM Cortex-M cores do not include dedicated floating-point units, 
the models should first be quantized before deployment.
CMSIS-NN~\cite{lai2018cmsis} is a library from ARM that helps quantizing and deploying NN models onto the ARM Cortex-M cores.
Specifically, the library leverages fixed-point quantization~\cite{lin2016fixed,jacob2018quantization,yao2020hawqv3} with power-of-two scaling factors so that quantization and dequantization processes can be carried out efficiently with bit shifting operations.
GAP-8~\cite{flamand2018gap}, a RISC-V SoC (System on Chip) for edge inference with a dedicated CNN accelerator, is another example of an edge processor that only supports integer arithmetic. 
While programmable general-purpose processors are widely adopted due to their flexibility, 
Google Edge TPU, a purpose-built ASIC chip, is another emerging solution for running inference at the edge.
Unlike Cloud TPUs that run in Google data centers with a large amount of computing resources, 
the Edge TPU is designed for small and low-power devices, and thereby it only supports 8-bit arithmetic.
NN models must be quantized using either quantization-aware training or post-training quantization of~TensorFlow.

Figure~\ref{fig:hw} plots the throughput of different commercial edge processors that are widely used for NN inference at the edge.
In the past few years, there has been a significant improvement in the computing power of the edge processors, 
and this allows deployment and inference of costly NN models that were previously available only on servers.
Quantization, combined with efficient low-precision logic and dedicated deep learning accelerators, has been one important driving force for the evolution of such edge processors.

While quantization is an indispensable technique for a lot of edge processors, 
it can also bring a remarkable improvement for non-edge processors, e.g., to meet
Service Level Agreement (SLA) requirements such as 99th percentile latency.
A good example is provided by the recent NVIDIA Turing GPUs, and in particular T4 GPUs, which include the Turing Tensor Cores.
Tensor Cores are specialized execution units designed for efficient low-precision matrix multiplications.

\section{Future Directions for Research in Quantization}
\label{sec:future}
Here, we briefly discuss several high level challenges and opportunities for future research in quantization.
This is broken down into quantization software, hardware and NN architecture co-design, coupled
compression methods, and quantized training.

\textbf{Quantization Software:}
With current methods, it is straightforward to quantize and deploy different NN models to INT8, without losing accuracy. There are several software packages
that can be used to deploy INT8 quantized models (e.g., Nvidia's TensorRT, TVM, etc.), each with
good documentation. Furthermore, the implementations are also quite optimal and one can easily
observe speed up with quantization.
However, the software for lower bit-precision quantization is not widely available, and sometimes
it is non-existent. For instance, Nvidia's TensorRT does not currently support sub-INT8 quantization.
Moreover, support for INT4 quantization was only recently added to TVM~\cite{yao2020hawqv3}.
Recent work has shown that low precision and mixed-precision quantization with
INT4/INT8 works in practice~\cite{yao2020hawqv3,wang2018haq,hubara2020improving,zhou2017adaptive,wang2018haq,dong2019hawq,van2020bayesian, yang2021bsq,Qu_2020_CVPR, wang2020apq, hu2021opq, ning2021simple, habi2020hmq, rusci2020leveraging}. Thus, developing efficient software APIs for lower precision quantization
will have an important impact.

\textbf{Hardware and NN Architecture Co-Design:}
As discussed above, an important difference between classical work in low-precision
quantization and the recent work in machine learning is the fact that NN parameters
may have very different quantized values but may still generalize similarly well.
For example, with quantization-aware training, we might converge to a different solution, far away from the original solution with single precision parameters, but
still get good accuracy. One can take advantage of this degree of freedom and also
adapt the NN architecture as it is being quantized.
For instance, the recent work of~\cite{chin2020one} shows that changing
the width of the NN architecture could reduce/remove generalization gap after quantization.
One line of future work is to adapt jointly other architecture parameters, such
as depth or individual kernels, as the model is being quantized.
Another line of future work is to extend this co-design to hardware architecture.
This may be particularly useful for FPGA deployment, as one can explore many different possible
hardware configurations (such as different micro-architectures of multiply-accumulate elements), and then couple this with the NN architecture and quantization co-design.

\textbf{Coupled Compression Methods:}
As discussed above, quantization is only one of the methods for efficient deployment of NNs.
Other methods include efficient NN architecture design, co-design of hardware and NN architecture, pruning, and knowledge distillation.
Quantization can be coupled with these other approaches.
However, there is currently very little work exploring what are the optimal combinations of these
methods. For instance, pruning and quantization can be applied together to a model to reduce
its overhead~\cite{hawks2021ps,liang2021pruning}, and it is important to understand the best combination of structured/unstructured pruning and quantization. Similarly, another future direction
is to study the coupling between these methods and other approaches described above.

\textbf{Quantized Training:}
Perhaps the most important use of quantization has been to accelerate NN training with half-precision~\cite{courbariaux2014training,gupta2015deep,ginsburg2017tensor,micikevicius2017mixed}. This has enabled the use of much 
faster and more power-efficient reduced-precision logic for training. However, it has been very difficult to push this further down to INT8
precision training.
While several interesting works exist in this area~\cite{johnson2018rethinking,mellempudi2019mixed,banner2018scalable,langroudi2019cheetah,cambier2020shifted}, the proposed methods often
require a lot of hyperparameter tuning, or they only work for a few NN models on relatively easy
learning tasks. 
The basic  problem is that, with INT8 precision, the training can
become unstable and diverge.
Addressing this challenge can
have a high impact on several applications, especially for training at the edge.

\section{Summary and Conclusions}
\label{sec:conclusions}

As soon as abstract mathematical computations were adapted to computation on digital computers, the problem of 
efficient representation, manipulation, and communication of the numerical values in those computations arose.  
Strongly related to the problem of numerical representation is the problem of quantization: in what manner should 
a set of continuous real-valued numbers be distributed over a fixed discrete set of numbers to minimize the number
of bits required and also to maximize the accuracy of the attendant computations? While these problems are as old 
as computer science, these problems are especially relevant to the design of efficient NN models.
There are several reasons for this. 
First, NNs are computationally intensive. So, the efficient representation of numerical values is 
particularly important. Second, most current NN models are heavily over-parameterized. So, there is ample 
opportunity for reducing the bit precision without impacting accuracy. Third, the layered structure of NN models 
offers an additional dimension to explore. Thus, different layers in the NN have different impact on the loss function, 
and this motivates interesting approaches such mixed-precision quantization. 

Moving from floating-point representations to low-precision
fixed integer values represented in eight/four bits or less holds the
potential to reduce the memory footprint and latency.
~\cite{lin2019automating} shows that INT8 inference of popular computer vision models, including ResNet50~\cite{he2016deep}, VGG-19~\cite{simonyan2014very}, and inceptionV3~\cite{szegedy2016rethinking} using TVM~\cite{chen2018tvm} quantization library, can achieve 3.89$\times$, 3.32$\times$, and 5.02$\times$ speedup on NVIDIA GTX 1080, respectively. 
\cite{salvator2019int4} further shows that 
INT4 inference of ResNet50 could bring an additional 50-60\% speedup on NVIDIA T4 and RTX, compared to its INT8 counterpart, emphasizing the importance of using lower-bit precision to maximize efficiency.
Recently,~\cite{yao2020hawqv3} leverages mix-precision quantization to achieve 23\% speedup for ResNet50, as compared to INT8 inference without accuracy degradation, 
and~\cite{kim2021bert} extends INT8-only inference to BERT model to enable up to 4.0$\times$ faster inference than FP32.
While the aforementioned works focus on acceleration on GPUs,~\cite{jain2020efficient} also obtained 2.35$\times$ and 1.40$\times$ latency speedup on Intel Cascade Lake CPU and Raspberry Pi4 (which are both non-GPU architectures), respectively, through INT8 quantization of various computer vision models.
As a result, as our bibliography attests, the problem of quantization in NN models has been a highly active research area. 

In 
this work, we have tried to bring some conceptual structure to these very diverse efforts. We began with a 
discussion of topics common to many applications of quantization, such as uniform, non-uniform, symmetric, 
asymmetric, static, and dynamic quantization.  We then considered quantization issues that are more unique to the 
quantization of NNs. These include layerwise, groupwise, channelwise, and sub-channelwise quantization. 
We further considered the inter-relationship between training and quantization, and we discussed the advantages 
and disadvantages of quantization-aware training as compared to post-training quantization. Further nuancing the 
discussion of the relationship between quantization and training is the issue of the availability of data. The 
extreme case of this is one in which the data used in training are, due to a variety of sensible reasons such as 
privacy, no longer available. This motivates the problem of zero-shot quantization. 

As we are particularly concerned about efficient NNs targeted for edge-deployment, we considered problems that are
unique to this environment. These include quantization techniques that result in parameters represented by fewer 
than 8 bits, perhaps as low as binary values. We also considered the problem of integer-only quantization, which 
enables the deployment of NNs on low-end microprocessors which often lack floating-point units.

With this survey and its organization, we hope to have presented a useful snapshot of the current research in quantization for Neural Networks and to have given an intelligent organization to ease the evaluation of future research in this area.

\section*{Acknowledgments}
The UC Berkeley team also acknowledges gracious support from Samsung (in particular Joseph Hassoun), Intel corporation, Intel VLAB team, Google TRC team, and Google Brain (in particular Prof. David Patterson, Dr. Ed Chi, and Jing Li).
Amir Gholami was supported through through funding from Samsung SAIT.
Our conclusions do not necessarily reflect the position or the policy of our sponsors, and no official endorsement should be inferred.

{
\bibliographystyle{plain}
\bibliography{ref.bib}
}


\end{document}